\documentclass[a4paper,fleqn]{elsarticle}

\usepackage[numbers]{natbib}
\usepackage{algorithm}  
\usepackage{algorithmic}
\usepackage{multirow}
\usepackage{arydshln}
\usepackage{graphicx} 
\usepackage{epstopdf}

\def\tsc#1{\csdef{#1}{\textsc{\lowercase{#1}}\xspace}}
\tsc{WGM}
\tsc{QE}
\tsc{EP}
\tsc{PMS}
\tsc{BEC}
\tsc{DE}

\begin{document}
\let\WriteBookmarks\relax
\def\floatpagepagefraction{1}
\def\textpagefraction{.001}

\shorttitle{PPL Learning with CRI Loss for Imbalanced Data Classification}

\shortauthors{Xu et~al.}

\title [mode = title]{Phased Progressive Learning with Coupling-Regulation-Imbalance Loss for Imbalanced Data Classification}                      



%
\author[1]{Liang Xu}[style=chinese,orcid=0000-0001-7841-106X]






\affiliation[1]{organization={Suzhou Institute for Advanced Research, University of Science and Technology of China},
    addressline={Suzhou 215123}, 
    country={China}}

\author[2]{Yi Cheng}[style=chinese]
\author[2]{Fan Zhang}[style=chinese]
\author[2]{Bingxuan Wu}[style=chinese]
\author[2]{Pengfei Shao}[style=chinese]
\author[1]{Peng Liu}[style=chinese]

\author[1]{Shuwei Shen}[style=chinese, orcid=0000-0002-6083-1442]
\cormark[1]
\cortext[cor1]{Corresponding author}
\ead{swshen@ustc.edu.cn}

\author[3]{Peng Yao}[style=chinese, orcid=0000-0003-0717-2836]
\cormark[1]
\ead{yaopeng@ustc.edu.cn}

\author[1]{Ronald X.Xu}[style=chinese, orcid=0000-0003-2486-5677]
\cormark[1]
\ead{xux@ustc.edu.cn}



\affiliation[2]{organization={Department of Precision Machinery and Precision Instrument, University of Science and Technology of China},
    city={Hefei 230026},
    country={China}}

\affiliation[3]{organization={School of Microelectronics, University of Science and Technology of China},
    city={Hefei 230026},
    country={China}}    






\begin{abstract}
Deep convolutional neural networks often perform poorly when faced with datasets that suffer from quantity imbalances and classification difficulties. Despite advances in the field, existing two-stage approaches still exhibit dataset bias or domain shift. To counter this, a phased progressive learning schedule has been proposed that gradually shifts the emphasis from representation learning to training the upper classifier. This approach is particularly beneficial for datasets with larger imbalances or fewer samples. Another new method a coupling-regulation-imbalance loss function is proposed, which combines three parts: a correction term, Focal loss, and LDAM loss. This loss is effective in addressing quantity imbalances and outliers, while regulating the focus of attention on samples with varying classification difficulties. These approaches have yielded satisfactory results on several benchmark datasets, including Imbalanced CIFAR10, Imbalanced CIFAR100, ImageNet-LT, and iNaturalist 2018, and can be easily generalized to other imbalanced classification models.
\end{abstract}





\maketitle

\section{Introduction}

Thanks to the noteworthy efforts of researchers, remarkable results have been achieved with deep convolutional neural networks (DCNN) for large-scale and uniformly distributed datasets \cite{he2016deep,Tan2019EfficientNetRM,Radosavovic_2020_CVPR}, such as ImageNet \cite{deng2009imagenet} and MS COCO \cite{lin2014microsoft}. However, in real scenarios, datasets generally have “imbalance” characteristic. Most of these imbalance problems are compounded by the following: 1) Quantity imbalance between different classes, wherein a few classes (a.k.a. head classes) occupy most of the data and most classes (a.k.a. tail classes) have rarely few samples \cite{zhou2020bbn,van2017devil}. 2) Classification difficulty imbalance. Samples in some head classes cannot be distinguished from similar samples in other head or tail classes. For example, the task of classifying skin lesions presents a significant challenge, particularly when distinguishing between melanoma and other skin conditions such as dermatofibromas and moles \cite{tschandl2018ham10000,kawahara2018seven}. Although melanoma is a more serious disease than the latter, these lesions often share similar morphologic characteristics and require careful examination and analysis to accurately differentiate. Furthermore, certain samples within the dataset, commonly referred to as outliers \cite{li2019gradient,yao2021single}, may be subject to issues such as pollution or a drastically imbalanced foreground-background ratio \cite{oksuz2020imbalance}. For example, some data augmentation methods, such as random cropping may introduce samples that contain only part or none of the foreground, resulting in large losses during convergence training. Thereafter, if the converged model is forced to learn to classify these outliers better, it tends to be less accurate in classifying many other examples \cite{li2019gradient}. Secondly, in the real scene, the problem of "imbalance" is often accompanied by the problem of insufficient samples, it will be difficult to collect enough data to train the model, which will lead to the problem of over-fitting caused by repeated training of the model with few samples \cite{park2021few,frikha2021few,liu2021relative}. It has been a challenging task to alleviate the two kinds of imbalance problems, the outlier problem and the problem of insufficient samples\cite{japkowicz2002class,buda2018systematic}.

Various strategies have been proposed to address the problem of quantity imbalance, with re-balancing methods being the most commonly employed, including one-stage methods and two-step approaches \cite{zhou2020bbn}. One-stage methods predominantly comprise the re-weighting (RW) method \cite{huang2016learning,wang2017learning} and re-sampling (RS) method \cite{buda2018systematic,shen2016relay}. Re-weighting prevents the network from ignoring rare classes by inverting the loss weighting factor for the number of categories. Re-sampling adjusts the distribution of training instances according to class size. The two-stage approaches divide the training process into two distinct stages. In Stage 1, the networks are trained as usual on the originally imbalanced data to initialize appropriate weights for deep layers’ features. In Stage 2, re-balancing is employed, and the networks are fine-tuned with a lower learning rate to facilitate the  optimization of the upper classifier of the DCNN. Although two-stage approaches perform better than one-stage methods, the abrupt transition between stages can result in dataset bias or domain shift \cite{luo2019taking,zhong2021improving}. For example, there is an inconsistency in the distribution of data that is sampled following different strategies in Stage 2 and Stage 1 \cite{zhong2021improving}. In addition to re-balancing methods, mixup methods \cite{zhang2017mixup,chou2020remix} have been demonstrated to be effective in improving the classification performance for imbalanced datasets. This technique involves creating new virtual samples with convex combination pairs of features and labels. The efficacy of the label-distribution-aware margin (LDAM) loss on quantity imbalance has been demonstrated \cite{cao2019learning}, encouraging the use of larger margins for tail classes.

To more effectively mitigate the dataset bias or domain shift that exists in the two-stage approaches more effectively, we propose a phased progressive learning (PPL) schedule. A progressive transition phase is inserted between the two stages of the two-stage approaches. It helps to realize a gradual and smooth training transition from the universal pattern of representation learning to the upper classifier training \cite{zhou2020bbn}. Moreover, the proposed PPL can work easily in combination with RW, RS, and mixup, forming phased progressive weighting (PPW), phased progressive sampling (PPS), and phased progressive mixup (PPmix) to solve imbalance problems more accurately. Surprisingly, we also found that progressive training using the PPL can effectively prevent the over-fitting problem caused by repeated training of small samples. 

The above studies have made remarkable progress in solving quantity imbalance problems \cite{wang2021contrastive,deng2021pml,wu2021adversarial}, while most of them ignore the problem of classification difficulty imbalance problem. Focal loss \cite{lin2017focal} is one of the few methods that addresses the problem of classification difficulty imbalance. It introduces a modulating term to the CE loss to improve the training results on samples with classification difficulty imbalance. To simultaneously address the problems of quantity imbalance and classification difficulty imbalance, we further propose a coupling-regulation-imbalance (CRI) loss function by coupling the Focal loss and the LDAM loss. The Focal loss part in the CRI loss allows to regulate the attention for samples of varying classification difficulties, and the LDAM loss part helps to solve quantity imbalance problems. A correction term is incorporated into the CRI loss to truncate possible huge losses, with the goal of reducing the influence of outliers on the DCNN training. 

The main contributions of this paper are as follows: (a) A three-stage PPL schedule with a progressive transition phase is proposed to facilitate a smoother transition from universal representation learning to classifier training. PPL outperforms other re-balancing methods on a variety of datasets, especially those with larger imbalances or of fewer samples. As a general training schedule, PPL can be easily combined with other methods for imbalanced classification tasks due to its simplicity and effectiveness. (b) A novel coupling-regulation-imbalance loss is proposed that includes a correction term, Focal loss, and LDAM loss. The loss can effectively deal with the quantity imbalance, regulate the focus-of-attention for samples with different classification difficulties and limit the resulting huge loss for outliers. (c) Achieve state-of-the-art classification results on all four imbalanced benchmark datasets when combined with PPL schedule and CRI loss, including Imbalanced CIFAR10 \cite{cui2019class}, Imbalanced CIFAR100 \cite{cui2019class}, ImageNet-LT \cite{liu2019large}, and iNaturalist 2018 \cite{van2018inaturalist}. All the source codes of our methods are available at \url{https://github.com/simonustc/Imbalance_PPL_CRI}.

\section{Related work}
\subsection{Re-weighting}
Re-weighting methods are widely used in imbalanced visual recognition and typically introduce a loss weighting factor into the loss function that is inversely proportional to the number of samples, and select the softmax cross-entropy (CE) loss function as the baseline: 
\begin{align}
\label{equ1}
\ \mathcal{L}_{RW}={-(\frac{1}{n_y})}log(p_y)
\end{align}
where $p_y=e^{z_y}/(\sum_{j=1}^{C}e^{z_j})$, $C$ is the total number of classes, $z_j$ is the predicted output for class j, $z_y$ is the predicted output for the ground truth class $y\in[1,2,\ldots,C]$, $n_y$ is the number of samples in class y.

However, if the dataset is extremely imbalanced, re-weighting may no longer contribute to model optimization \cite{cao2019learning}. Because the weights are concentrated in the tail classes, the network is more sensitive to fluctuations in the fit of the tail classes, which greatly increases the model variance \cite{wang2020long}. Cui et al. \cite{cui2019class} proposed the concept of effective number, arguing that each sample represents an area covering the feature space rather than a single point. Subsequently, the class-balanced (CB) method was proposed as a way to re-weight the samples using their inverted effective number instead of the actual number. According to the theory of effective numbers, the CB loss with softmax CE loss is updated as follows:
\begin{align}
\label{equ2}
\ \mathcal{L}_{CB}={-(\frac{1-\beta}{1-\beta^{n_y}})}log(p_y)
\end{align}
where ${(1-\beta)}/(1-\beta^{n_y})$ represents the inverse of the effective number of samples and $\beta$ is a hyperparameter. 

On the other hand, hinge loss, including Large-Margin Softmax \cite{liu2016large}, Additive Margin Softmax \cite{wang2018additive}, helps the classifier expand the interclass boundary by aiming to obtain the "maximum margins". Cao et al. \cite{cao2019learning}  derived a theoretical formulation by exploring the margins of training examples and designed a label-distribution-aware margin (LDAM) loss to encourage larger margins for the tail classes. The LDAM loss redefines $p_y=(e^{z_y-\Delta_{y}})/(e^{z_y-\Delta_{y}}+\sum_{j\notin y}e^{z_j})$ in the CE loss and is shown as follows:
\begin{align}
\label{equ3}
\ \mathcal{L}_{LDAM}=-log\frac{e^{z_y-\Delta_{y}}}{e^{z_y-\Delta_{y}}+\sum_{j\notin y}e^{z_j}}
\end{align}
where $\Delta_{y}=s/n_y^{1/4}$ and $S$ is a hyperparameter. For the tail classes, the value of $n_y$ is small while $\Delta_{y}$ becomes quite large, causing the tail classes to expand outward, improving their classification performance.

In addition, there are also studies that assign weights to the samples based on their other characteristics. For example, Focal loss \cite{lin2017focal} is proposed based on CE loss by introducing a modulation factor:
\begin{align}
\label{equ4}
\ \mathcal{L}_{Focal}=-(1-p_y)^{\gamma}log(p_y)
\end{align}
where $\gamma$ is a hyperparameter and the Focal loss is equivalent to the CE loss when $\gamma=0$. As $\gamma$ increases, the Focal loss facilitates training to focus more on the difficult samples, leading to a more balanced performance.

\begin{figure*}
\centering
\includegraphics[scale=0.8]{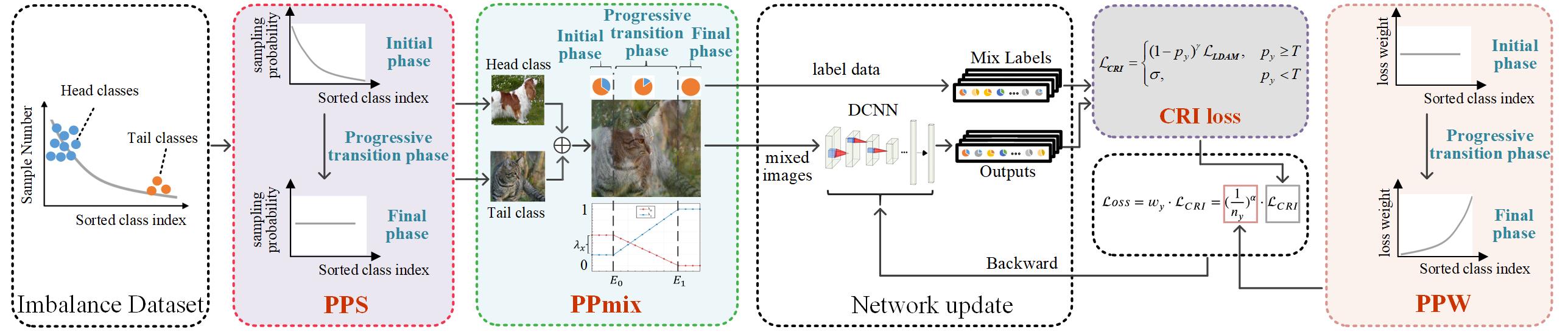}
\caption{The flowchart of a demo training framework for a DCNN, which incorporates our proposed PPL methods (PPS, PPmix, and PPW) and CRI loss. It is noteworthy that these methods can be combined with each other or in pairs as illustrated in the figure, and they can also be used as separate modules in conjunction with traditional methods.}
\label{fig1}
\end{figure*}

\subsection{Re-sampling}
Re-sampling is another prominent preprocessing technique, and it helps to obtain balanced training data either by resampling the originally imbalanced data or by generating new data.

Re-sampling methods can be divided into two groups: over-sampling \cite{buda2018systematic,shen2016relay} and under-sampling \cite{buda2018systematic,he2009learning}, which achieve sample balance by increasing the number of samples in the tail class or decreasing the number of samples in the head class during the training phase. Despite their considerable advantages, over-sampling can lead to over-fitting of the tail classes, and under-sampling discards a significant amount of useful data \cite{zhou2020bbn}.

To achieve more efficient re-sampling, Kang et al. \cite{kang2019decoupling} proposed a class-balanced (C-Balance) sampling method, as shown in (\ref{equ5}):
\begin{align}
\label{equ5}
\ p_j=\frac{n_j^q}{\sum_{i=1}^{C}n_i^q}
\end{align}
where $p_j$ is the probability of selecting a sample from class $j$. $q$ is a hyperparameter, and changing $q$ indicates differing re-sampling strategies. If $q=0$ in C-Balance sampling, then the probability $p_j^{CB}=1/(\sum_{i=1}^C)=1/C$, resulting in equal probability sampling in each class. When $q$ is set as $1/2$, then (\ref{equ5}) becomes Square-root sampling \cite{kang2019decoupling,zhang2021bag}. When $q$ is set to 1, the probability of selecting samples is equal to the inverse of the total number in the corresponding class, and (\ref{equ5}) reverts to random sampling.

In addition to data replication, another effective strategy for over-sampling is to generate synthetic data for the tail classes. Chawla et al. \cite{chawla2002smote} proposed a synthetic minority over-sampling technique (SMOTE), where SMOTE finds the k-nearest neighbors for each tail class sample, and draws a random neighborhood is drawn. The drawn features are then linearly combined with features along the tail classes to generate a virtual sample. The formula for generating samples $\widetilde x$ using SMOTE is as follows:
\begin{align}
\label{equ6}
\ \widetilde x=x+(\widetilde x-x)*r
\end{align}
where $x$ represents the tail class sample, $\widetilde x$ represents the field selected by sample $x$, and $r$ represents a random number uniformly distributed from $[0,1]$. In addition, many other SMOTE-based methods have also been developed, including borderline-SMOTE \cite{han2005borderline}, safe-level-SMOTE \cite{bunkhumpornpat2009safe}, and MBS \cite{liu2019model}, etc.

\subsection{Two-stage approaches}
Cao et al. \cite{cao2019learning} first proposed the two-stage deferred RW (DRW) and deferred RS (DRS) methods. It routinely trains in a regular pattern for Stage 1, then anneals the learning rate and trains with re-balancing methods in Stage 2. Here, the learning in Stage 1 provides a good initialization for the training in Stage 2. 

Kang et al. \cite{kang2019decoupling} divided the training process into representation learning and classifier learning, which correspond to the first stage and the second stage, respectively. Note that, the weights of the feature layers are fixed and only the classifier is fine-tuned in Stage 2. Zhou et al. \cite{zhou2020bbn} proposed a bilateral branch network (BBN) to combine representation learning and classifier rebalancing. It stimulates the DRS process by dynamically combining instance samplers and reverse samplers, and adjusts the bilateral branches using the cumulative learning strategy. 

Another common approach is progressively-balanced (P-B) sampling \cite{kang2019decoupling,zhang2021bag}, where the transition from random sampling to C-Balance sampling is implemented throughout the entire training process. The probability $p_j^{PB}$ of P-B is given by (\ref{equ7}):
\begin{align}
\label{equ7}
\ p_j^{PB}(E)=(1-\frac{E}{E_t})p_j+\frac{E}{E_t}p_j^{PB}
\end{align}
where $E_t$ is the total number of epochs, and $E$ represents the current training epoch. 

However, two-stage approaches cannot avoid the problems that may cause dataset bias or domain shift when abrupt transitions between stages \cite{luo2019taking,zhong2021improving}.

\subsection{Regularization}
According to Byrd et al. \cite{byrd2019effect}, the effectiveness of re-weighting may be insufficient when no regularization is applied. Then, regularization methods such as Mix up \cite{zhang2017mixup} are proposed, which improve the generalization of DCNN by linearly combining arbitrary pairs of samples in the dataset. It is implemented as shown in (\ref{equ8}) and (\ref{equ9}) by using a mixing factor $\lambda$, which is sampled from the beta distribution:
\begin{align}
\label{equ8}
\ \widetilde x=\lambda{x_1}+(1-\lambda)x_2
\end{align}
\begin{align}
\label{equ9}
\ \widetilde y=\lambda{y_1}+(1-\lambda)y_2
\end{align}
where each newly mixed sample $(\widetilde x, \widetilde y)$ is generated through a combination of an arbitrary sample pair $(x_1, y_1)$ and $(x_2, y_2)$. y represents the label of sample x. Another approach, Manifold Mixup \cite{verma2019manifold}, combines the features linearly in the embedding space instead of mixing samples directly. The operation is performed by randomly combining the features at layer k of the network. In addition, mixup shifted label-aware smoothing (MisLAS) \cite{zhong2021improving} combines mixup and label-aware smoothing to improve calibration and performance.

Chou et al. \cite{chou2020remix} then introduced Remix, where labels are more appropriate for a few classes and  are created by relaxing the mixing factor. It performs linear interpolation weighting by relaxing the mixing factor, thus updating (\ref{equ8}) and (\ref{equ9})  as follows:
\begin{align}
\label{equ10}
\ \widetilde x=\lambda_{x}{x_1}+(1-\lambda_{x}){x_2}
\end{align}
\begin{align}
\label{equ11}
\ \widetilde y=\lambda_{y}{y_1}+(1-\lambda_{y}){y_2}
\end{align}
where Remix transforms $\lambda$ into $\lambda_x$ and $\lambda_y$ in the Mix up method \cite{zhang2017mixup}. $\lambda_x$ is an image mixing factor that is randomly chosen from the $\beta$  distributed values and $\lambda_y$ is a label mixing factor, which is defined as:
\begin{align}
\label{equ12}
\lambda_y=
\begin{cases}
\lambda_0,  & n_1/n_2\geq \kappa \quad and\quad  \lambda_x $\textless$ \tau \\
\lambda_1, & n_1/n_2 \leq 1/\kappa \quad and\quad  \lambda_x $\textgreater$ 1-\tau \\
\lambda_x, & otherwise
\end{cases}
\end{align}
where $\kappa$ and $\tau$ are two hyperparameters in the Remix method \cite{chou2020remix}. $n_1$ and $n_2$ denote the number of samples in the class of sample 1 and sample 2, respectively. $\lambda_0$ and $\lambda_1$ are fixed to 0 and 1.

Unlike other hybrid methods, the Remix method improves the performance of models on imbalanced classification tasks by modifying $\lambda_y$ to skew the model toward the tail end of the distribution. However, the skewing toward the tail end from the start of training, like other re-sampling methods, may result in excessive bias toward the tail end, which in turn is detrimental to the head classes. Additionally, it is not conducive to the learning of universal features.

In addition to mixup-based approaches, the Knowledge Distillation (KD) method in regularization has also been utilized for addressing class imbalance. KD was originally proposed by Hinton \cite{hinton2015distilling} and compresses knowledge into a compact student network by training the student network to mimic the behavior of the teacher network. The techniques of Learning from multiple experts (LFME) \cite{xiang2020learning} and routing diverse distribution-aware experts (RIDE) \cite{wang2020long} aim to distill a variety of networks into a single, unified model that can be used effectively for imbalanced datasets.  

\begin{figure}
\centering
\includegraphics[scale=.65]{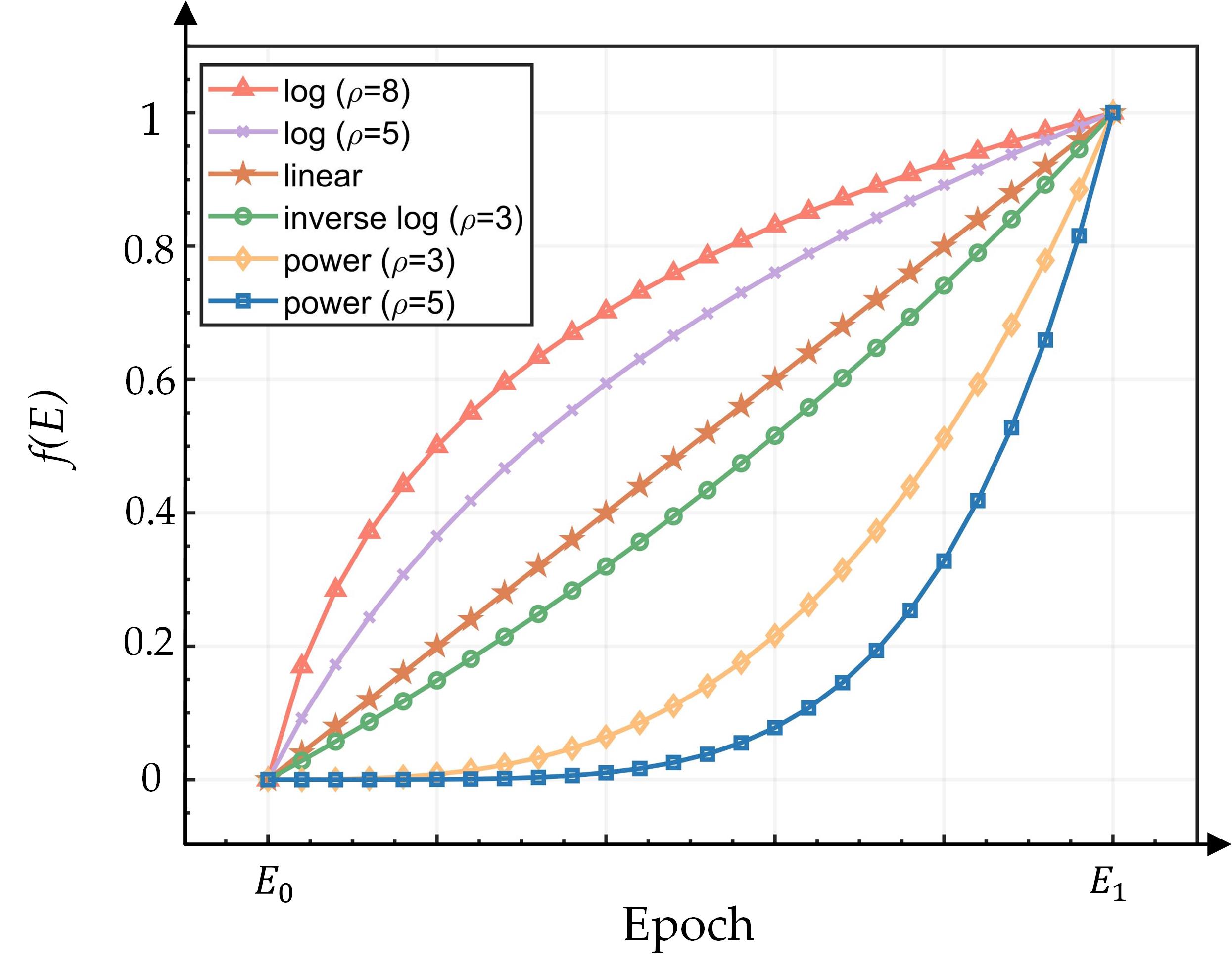}
\caption{Basic forms of different transformation functions in PPL, including log form, inverse log form, and power-law form with different $\rho$.}
\label{fig2}
\end{figure}

\begin{algorithm*}
\caption{Phased Progressive Weighting}
\fontsize{9.0pt}{9.0pt}\selectfont
\begin{algorithmic}[1]
\renewcommand{\baselinestretch}{5}
\label{algorithm1}
\REQUIRE Dataset $\mathcal D={(x_i,y_i)}_{i=1}^{n}$. A parameterized model $\mathbb M_\theta$; $f(E)$ $\leftarrow$ transformation function\
\STATE initialize the model parameters $\theta$ randomly\ 

\FOR{epoch $E=0 \ to \ E_0$} 
\STATE $\mathcal B$ $\leftarrow$ SampleMiniBatch$(\mathcal D,m)$ \hfill $\triangleright$ a mini-batch of $m$ examples \\
\STATE $\mathcal L(\mathbb M_\theta) \leftarrow \frac{1}{m}\sum\nolimits_{(x,y) \in \mathcal B} \mathcal L((x,y);\mathbb M_\theta)$ \hfill $\triangleright$ during the first phase \\
\STATE $\mathbb M_\theta \leftarrow \mathbb M_\theta - \omega \nabla_\theta \mathcal L(\mathbb M_\theta)$ \hfill $\triangleright$ one SGD step with learning rate $\omega$ \\
\ENDFOR
\FOR{epoch $E=E_0 \ to \ E_1$}
\STATE $\mathcal B$ $\leftarrow$ SampleMiniBatch$(\mathcal D,m)$ \hfill $\triangleright$ a mini-batch of $m$ examples \\
\STATE $\mathcal L(\mathbb M_\theta) \leftarrow \frac{1}{m}\sum\nolimits_{(x,y) \in \mathcal B} n_y^{-(\delta \cdot f(E))} \cdot \mathcal L((x,y);\mathbb M_\theta)$ \hfill $\triangleright$ during the progressive transition phase \\
\STATE $\mathbb M_\theta \leftarrow \mathbb M_\theta - \omega \frac{1}{\sum\nolimits_{(x,y) \in \mathcal B} n_y^{-(\delta \cdot f(E))}} \nabla_\theta \mathcal L(\mathbb M_\theta)$ \hfill $\triangleright$  SGD with re-normalized $\omega$ \\
\STATE Optional$:\omega \leftarrow \omega/ \rho $ \hfill $\triangleright$ anneal $\omega$ by a progressive hyperparameter $\rho$ \\
\ENDFOR
\FOR{epoch $E=E_1 \ to \ E_{t}$}
\STATE $\mathcal B$ $\leftarrow$ SampleMiniBatch$(\mathcal D,m)$ \hfill $\triangleright$ a mini-batch of $m$ examples \\
\STATE $\mathcal L(\mathbb M_\theta) \leftarrow \frac{1}{m}\sum\nolimits_{(x,y) \in \mathcal B} n_y^{-\delta } \cdot \mathcal L((x,y);\mathbb M_\theta)$ \hfill $\triangleright$ during the last phase \\
\STATE $\mathbb M_\theta \leftarrow \mathbb M_\theta - \omega \frac{1}{\sum\nolimits_{(x,y) \in \mathcal B} n_y^{-\delta}} \nabla_\theta \mathcal L(\mathbb M_\theta)$ \hfill $\triangleright$ SGD with re-normalized $\omega$ \\
\ENDFOR
\end{algorithmic}
\end{algorithm*}

\section{Phased progressive learning schedule}
In this study, we propose a phased progressive learning (PPL) schedule, where the entire training process is updated into three phases by introducing a progressive transition phase. The three phases are classified based on the phased training epoch threshold $[E_0, E_1]$, where the hyperparameters of $E_0$ and $E_1$ represent the start and end epochs of the progressive transition phase, respectively. During the initial phase $(E<E_0)$, the original imbalanced data is used to initialize the good weights for the feature layers (deep features, such as the features in the underlying convolutional layer). During this phase, the model undergoes the learning process and gradually reduces the loss to a minimum value. This phase is crucial for setting the appropriate weights for the feature layers, including the convolutional layer, so that the model can effectively extract and understand the relevant information from the input data. Combined with the non-convexity of the loss function, the weights of the depth feature are slightly optimized during the progressive transition phase $(E_0\leq E\leq E_1)$, rather than undergoing large changes. At the same time, the focus of training gradually shifts from representation learning to the upper classification layer of the model (i.e., an upper classifier, such as the upper fully connected layer). During the final phase $(E>E_1)$, the re-balancing methods are fully implemented to train the upper classifier. As introduced earlier, PPL smoothly connects the initial and final training phases via a progressive transition phase. As a result, PPL addresses the problem of dataset bias or domain shift caused by a sudden change in data or loss function in two-stage approaches. Secondly, through the follow-up experimental results, we also found that, as shown in Figure. \ref{fig4} (d, f), gradually training the network through the PPL method is also effective for over-fitting caused by repeated training on data sets with few samples.

Our proposed PPL  can be easily combined with other methods for address class imbalance problems, resulting in practical and concrete approaches. For example, PPW, PPS, and PPmix have been proposed by integrating PPL with re-weighting, re-sampling, and mixup, respectively. It should be noted that these methods can not only serve as standalone modules integrated into the training process of traditional DCNN, but can also be flexibly combined with each other or used in pairs. The flowchart shown in Figure. \ref{fig1} is a demo of a training framework for a DCNN that combines PPS, PPmix, and PPW, and introduces the CRI loss module. The PPS module is used to sample the imbalanced dataset, and the PPmix module is used to obtain mixed samples and their corresponding labels. Then, the DCNN performs forward propagation and the CRI loss module calculates the loss. Meanwhile, the PPW module modifies the weighting factors of the loss during its calculation. After the loss is calculated, the model parameters of the DCNN are updated by backward propagation. This iterative process is repeated until the training is complete. The following sections describe PPW, PPS, and PPmix in detail. 

\subsection{Phased progressive weighting}
According to (\ref{equ1}), the loss weighting factor of the phased progressive weighting (PPW) method is modified to (\ref{equ13}):
\begin{align}
\label{equ13}
\ w_i=(\frac{1}{n_i})^\alpha
\end{align}
where $n_i$ is the number of samples in the class i, and the total number of samples is $n=\sum_{i=1}^C{n_i}$. $\alpha$ is a parameter that varies with the training epoch $E$, and it is updated as follows: 
\begin{align}
\label{equ14}
\alpha =
\begin{cases}
0,  & E $\textless$ E_{0} \\
\delta\cdot f(E), & E_{0}\leq E\leq E_{1} \\
\delta, & E $\textgreater$ E_{1}
\end{cases}
\end{align}
where $\delta$ is a constant greater than 0. The diversity of weights can be further improved by setting a specific $\delta$. $f(E)$ is a monotonically increasing transformation function varying with $E$ that satisfies $f(E_0)=0$ and $f(E_1)=1$.

As seen in (\ref{equ14}), during the initial phase of representation learning, each class has the same loss weighting factor $(\alpha=0, w_y=1)$. In the progressive transition phase, $\alpha$ varies smoothly and continuously following the transformation function $f(E)$ from 0 to $\delta$. Similarly, during the final phase, the weights are set as values inversely proportional to the number of samples for each class $(\alpha=\delta,w_y=(1/n_y)^\delta)$, thus reflecting the relative importance of each class.

Note that the transformation function $f(E)$ can be concave or convex, as shown below, to accommodate different imbalance situations:

-Power-law form: $f(E)_{power}=(\frac{E-E_{0}}{E_{1}-E_{0}})^{\rho}$ ($\rho=1$ represents the linear form)

-Log form: $f(E)_{log}=\log_{\rho}[1+(\rho-1)\cdot(\frac{E-E_{0}}{E_{1}-E_{0}})]$ ($\rho=e$ represents the natural log form)

-Inverse log form: $f(E)_{in \_ log}=\rho^{[\frac{E-E_{0}}{E{1}-E_{0}}\cdot \log_{\rho}2]}-1$ ($\rho=e$ represents the natural inverse log form), where $\rho$ is a progressive hyperparameter for the further expansion of the smooth trend of the training form (Figure. \ref{fig2}).

The workflow of PPW is illustrated as a demo in Algorithm \ref{algorithm1}. The associated symbols are defined as follows: $x_i$ represents any sample with the label $y_i$ and $E_t$ is the last epoch in the whole training process.

\subsection{Phased progressive sampling}
The probability $p_j$ of sampling a data point of class j in the RS method is given by (\ref{equ5}). Unlike most RS methods \cite{kang2019decoupling} where $q$ is fixed, the phased progressive sampling (PPS) method in this paper dynamically updates $q$ as follows (\ref{equ15}):
\begin{align}
\label{equ15}
\ q =
\begin{cases}
1,  & E $\textless$ E_{0} \\
1-\delta\cdot f(E), & E_{0}\leq E\leq E_{1} \\
1-\delta, & E $\textgreater$ E_{1}
\end{cases}
\end{align}

The training is also divided into three phases and uses the same transformation function $f(E)$ as defined in the PPW. During the initial phase, $q=1$ means that the algorithm randomly selects from each class with equal probability. During the progressive transition phase, $f(E)$ is used to smooth the transition of $q$ from 1 to $1-\delta$. During the final phase, a hyperparameter $\delta$ is introduced to narrow the difference between the head and tail classes. In general, each class has an equal chance of being selected when $\delta$ is set to 1 $(q=0)$.

It should be noted that the progressively-balanced (P-B) sampling method \cite{kang2019decoupling} is similar to the progressive transition phase of the PPS, but it lacks the initial and final training phases. However, the initial phase is considered essential because training in the universal pattern on the original data can better initialize model parameters for subsequent training stages. During the equally important final phase, the training shifts completely to the balanced mode. In this situation, the training does not end immediately, but continues for a certain number of epochs. This strategy is conducive to the continuous updating of the upper classifier, which better matches the tail classes.

\begin{figure}
\centering
\includegraphics[scale=.65]{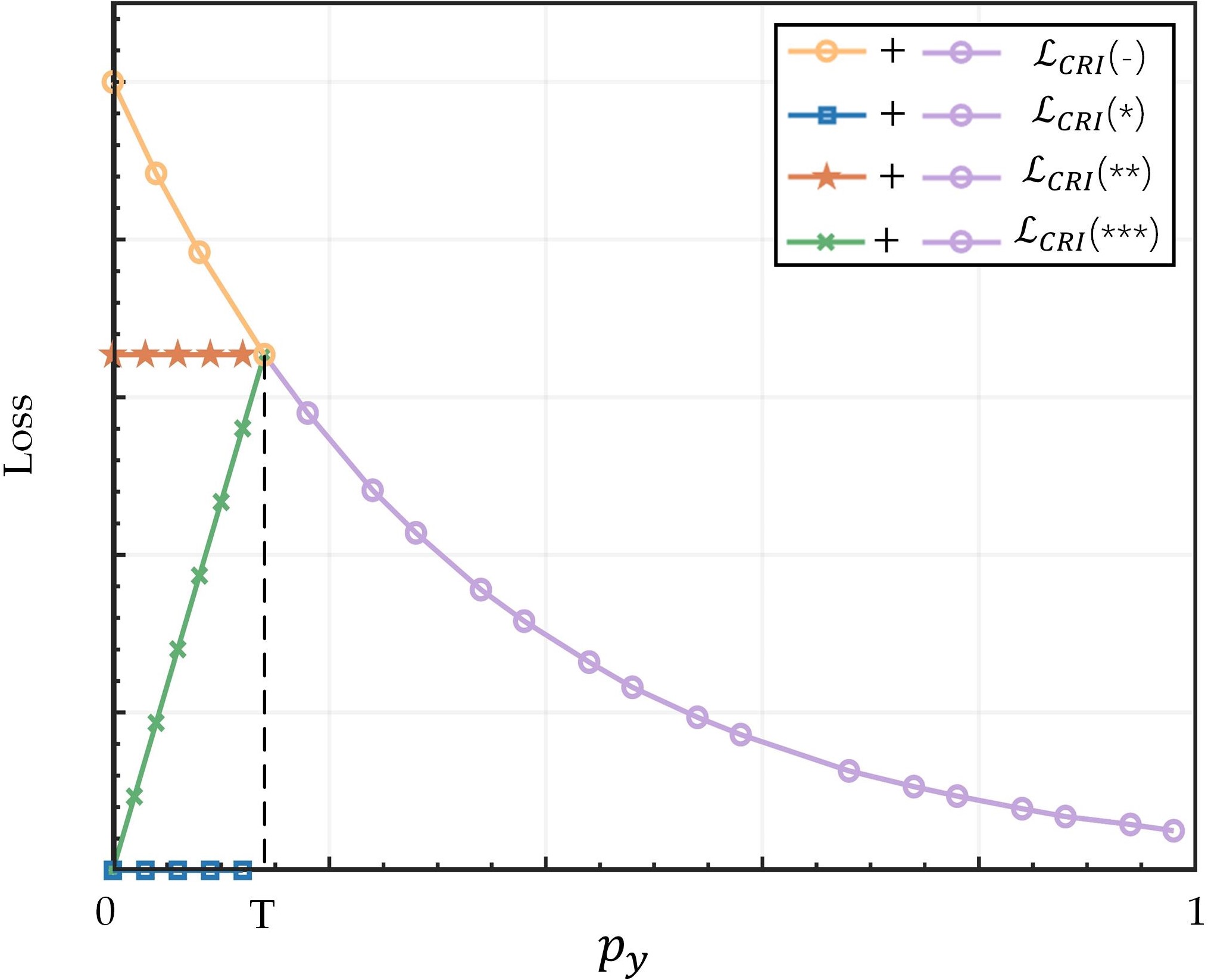}
\caption{The curve of CRI loss versus confidence in the correct class. $-$ denotes the loss curve without $\sigma$; $*$ denotes $\sigma=0$; $**$ denotes $\sigma=-(1-T)^\gamma logT$; $***$ denotes $\sigma=-(p_y/T)(1-T)^\gamma logT$.}
\label{fig3}
\end{figure}

\begin{table*}
\renewcommand\arraystretch{2}
\fontsize{9.0pt}{9.0pt}\selectfont
\centering
\caption{Setting 1: experimental setting of CRI+PPL. Setting 2: experimental setting of CRI+PPL+RIDE (applied to the routing diverse distribution-aware experts). LR: initial learning rate. LRS: learning rate schedule, the LR decay epoch interval and frequency. Epoch: the total number of epochs for training. BS: batch size. PPW, PPS, and PPmix: the progressive hyperparameters $\rho$ and the phased hyperparameter thresholds $(E_0, E_1)$ of PPW/S, and PPmix.}
\label{table1}
\begin{tabular}{cc:l:cc:c:c:lc:cc} 
\hline
\multicolumn{2}{c:}{Dataset}                  & LR  & \multicolumn{2}{c:}{LRS} & Epoch & BS  & \multicolumn{2}{c:}{PPW/S ($\rho$, [$E_0$, $E_1$])} & \multicolumn{2}{c}{PPmix ($\rho$, [$E_0$, $E_1$])}  \\ 
\hline\hline
\multirow{3}{*}{Setting 1} & Imbalanced CIFAR & 0.1 & {[}160,180] & 0.1        & 200   & 128 & - & -                                  & 5 & {[}100,160]                          \\
                           & ImageNet-LT      & 0.1 & cosine      & 0.1        & 200   & 256 & 5 & {[}100,160]                        & 5 & {[}100,160]                          \\
                           & iNaturalist 2018 & 0.1 & cosine      & 0.1        & 200   & 640 & 5 & {[}100,160]                        & 5 & {[}100,160]                          \\ 
\hline\hline
\multirow{3}{*}{Setting 2} & Imbalanced CIFAR & 0.1 & {[}120,160] & 0.01       & 200   & 128 & 5 & {[}100,160]                        & - & -                                    \\
                           & ImageNet-LT      & 0.1 & {[}60,80]   & 0.1        & 100   & 256 & 5 & {[}50,80]                          & - & -                                    \\
                           & iNaturalist 2018 & 0.2 & {[}60,80]   & 0.1        & 100   & 640 & 5 & {[}50,80]                          & - & -                                    \\
\hline
\end{tabular}
\end{table*}

\subsection{Phased progressive mixup}
The previously proposed mixup mitigates adversarial perturbations by increasing the diversity of the samples, and it has been shown to be effective when used in combination with re-balancing methods \cite{zhong2021improving,chou2020remix}. 

As shown in (\ref{equ12}), in the Remix method, $\lambda_0$ and $\lambda_1$ are fixed to 0 and 1, respectively. As a result, the decision boundary will be overly biased in favor of the tail classes, which will affect the overall recognition accuracy. To solve this problem, the phased progressive mixup (PPmix) method is proposed, as shown in (Figure. \ref{fig1}). PPmix combines PPL and Remix, where $\lambda_0$ and $\lambda_1$ in (\ref{equ16}) and (\ref{equ17}) are modified as follows:
\begin{align}
\label{equ16}
\lambda_0 =
\begin{cases}
\lambda_x,  & E $\textless$ E_{0} \\
\lambda_x(1-f(E)), & E_{0}\leq E\leq E_{1} \\
0, & E $\textgreater$ E_{1}
\end{cases}
\end{align}
\begin{align}
\label{equ17}
\lambda_1 =
\begin{cases}
\lambda_x,  & E $\textless$ E_{0} \\
\lambda_x(1-f(E))+f(E), & E_{0}\leq E\leq E_{1} \\
1, & E $\textgreater$ E_{1}
\end{cases}
\end{align}
where $f(E)$ is the transformation function, similar to PPW and PPS.

PPmix also divides the whole training process into three phases. During the initial phase, $\lambda_0=\lambda_1=\lambda_x$, and the training is in a universal pattern. During the progressive transition phase, as $E$ is updated, $\lambda_0$ transitions smoothly from $\lambda_x$ to 0 following $f(E)$. Similarly, $\lambda_1$ changes from $\lambda_x$ to 1. During the final phase, $\lambda_0$ is set to 0 and $\lambda_1$ is set to 1, where the algorithm marks more synthetic samples as tail classes. PPMix moves the decision boundary gradually, rather than doing so instantaneously, by creating new data points. The gradual relaxation of the mixing factors also helps the model focus training on the tail classes during the final phase.

\begin{figure*}
\centering
\includegraphics[scale=0.9]{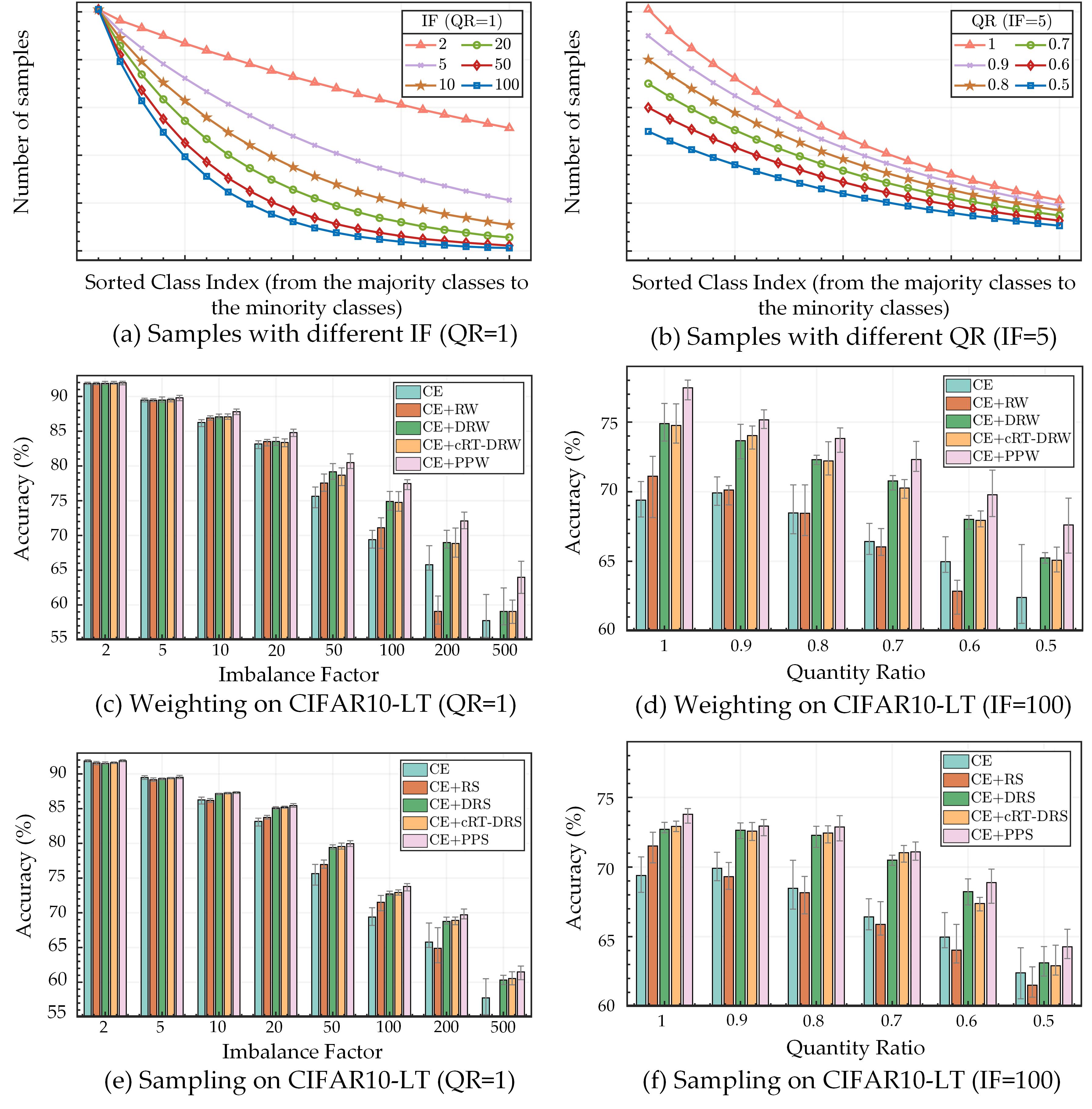}
\caption{(a-b) Data distribution in the imbalanced long-tailed datasets. (a) Number of training samples with different imbalance factor (IF). (b) Number of training samples with different quantity ratio (QR). (c-f) The average performance of models using cross-entropy (CE) loss, one-stage methods (RW, RS), two-stage approaches (DRW, DRS, and cRT), and the methods used in this paper (PPW and PPS) in training was repeated ten times. (c-d) Accuracy diagram for different IF and QR of weighting methods on CIFAR10-LT. (e-f) Accuracy diagram for different IF and QR of sampling methods on CIFAR10-LT-IF100 (for better display, values below the minimum ordinate are ignored).}
\label{fig4}
\end{figure*}

\section{Coupling-regulation-imbalance loss}
In addition to training strategies, we also focus on loss functions, which are equally important in dealing with imbalance problems. Since the LDAM loss works well for the problem of quantity imbalance, Focal loss focuses on dealing with the problem of classification difficulty imbalance. It is believed that $(1-p_y)^\gamma \mathcal{L}_{LDAM}$ integrating Focal loss and LDAM loss can more effectively deal with imbalance problems. At the same time, when $p_y$ of an outlier $\rightarrow$ 0, the loss $\rightarrow \infty$, which seriously misleads the optimization of network training. Therefore, the coupling-regulation-imbalance (CRI) loss is proposed by further introducing of a correction term to reduce the outlier interference: 
\begin{align}
\label{equ18}
\mathcal L_{CRI} =
\begin{cases}
(1-p_y)^\gamma \mathcal L_{LDAM},  & p_y \geq T \\
\sigma,  & p_y $\textless$ T
\end{cases}
\end{align}
where $T$ is a hyperparameter threshold and $\sigma$ is a correction term. Here $\sigma$ could be set to three values: $\sigma=0$, $\sigma=-(1-T)^\gamma logT$ and $\sigma=-(p_y/T)(1-T)^\gamma logT$. As shown in (Figure. \ref{fig3}), when the loss value is large enough $(p_y<T)$, there is an increasing likelihood of encountering an outlier. Therefore, the loss can be corrected to 0, a fixed value, or linearly decrease as a means of reducing outlier influence. 

Our proposed PPL method improves the classification performance of imbalanced datasets in terms of the training strategy. It can also be effectively combined with our CRI loss forming new methods, such as CRI+PPW, or even CRI+PPW+PPmix, etc. In addition, multi-expert methods, such as the RIDE method \cite{wang2020long}, also demonstrate their superior performance in classification tasks on imbalanced datasets. Therefore, we replace the LDAM loss in RIDE with the CRI loss, and introduce PPL, resulting in CRI+PPW+RIDE. Meanwhile, the original multi-expert module of RIDE is retained for its potential to reduce the variance of the model. Compared to the original RIDE, the method used in this study eliminates the routine module during the second stage, which means that the training becomes end-to-end and low-time-cost. 

\begin{table*}
\renewcommand\arraystretch{2}
\fontsize{9.0pt}{9.0pt}\selectfont
\centering
\caption{Top-1 accuracy $(\%)$ of various methods on CIFAR10-LT with different IF and QR (maximum performance in training repeated ten times).}
\label{table2}
\begin{tabular}{c:c:c:cccc:cccc} 
\hline
\multicolumn{1}{c}{}                     &     &      & \multicolumn{4}{c:}{Weighting}                       & \multicolumn{4}{c}{Sampling}          \\ 
\hline\hline
\multicolumn{1}{c}{Methods}              &     & CE   & RW   & DRW           & cRT-RW        & PPW           & RS   & DRS  & cRT-RS & PPS            \\ 
\hline\hline
\multirow{8}{*}{Imbalance Factor (QR=1)} & 2   & 92.1 & 92.0 & \textbf{92.2} & \textbf{92.2} & \textbf{92.2} & 91.8 & 91.7 & 91.7   & \textbf{92.1}  \\
                                         & 5   & 89.7 & 89.7 & 89.9          & 89.7          & \textbf{90.2} & 89.4 & 89.4 & 89.5   & \textbf{89.8}  \\
                                         & 10  & 86.7 & 87.2 & 87.5          & 87.5          & \textbf{88.2} & 86.5 & 87.2 & 87.3   & \textbf{87.5}  \\
                                         & 20  & 83.6 & 83.8 & 84.1          & 83.9          & \textbf{85.3} & 84.0 & 85.3 & 85.4   & \textbf{85.7}  \\
                                         & 50  & 77.0 & 78.8 & 80.3          & 79.7          & \textbf{81.7} & 77.6 & 79.8 & 80.0   & \textbf{80.4}  \\
                                         & 100 & 70.7 & 72.5 & 76.3          & 76.3          & \textbf{78.0} & 72.5 & 73.2 & 73.3   & \textbf{74.2}  \\
                                         & 200 & 68.5 & 61.3 & 70.7          & 71.0          & \textbf{73.4} & 67.9 & 69.4 & 69.4   & \textbf{70.6}  \\
                                         & 500 & 61.5 & 53.4 & 62.5          & 60.7          & \textbf{66.3} & 54.8 & 61.0 & 61.5   & \textbf{62.3}  \\ 
\hline\hline
\multirow{6}{*}{Quantity Ratio (IF=100)} & 1   & 70.7 & 72.5 & 76.3          & 76.3          & \textbf{78.0} & 72.5 & 73.2 & 73.3   & \textbf{74.2}  \\
                                         & 0.9 & 71.1 & 70.4 & 74.8          & 74.7          & \textbf{75.9} & 72.6 & 73.1 & 73.2   & \textbf{73.4}  \\
                                         & 0.8 & 70.5 & 70.5 & 72.6          & 73.6          & \textbf{74.6} & 70.3 & 72.9 & 73.0   & \textbf{73.7}  \\
                                         & 0.7 & 67.7 & 67.4 & 71.2          & 70.9          & \textbf{73.6} & 69.3 & 70.9 & 71.6   & \textbf{71.8}  \\
                                         & 0.6 & 66.8 & 63.6 & 68.3          & 68.6          & \textbf{71.6} & 65.9 & 69.2 & 67.8   & \textbf{69.8}  \\
                                         & 0.5 & 64.2 & 56.9 & 65.6          & 66.0          & \textbf{69.5} & 62.8 & 64.3 & 64.4   & \textbf{65.5}  \\
\hline
\end{tabular}
\end{table*}

\section{Experiments}
\subsection{Datasets}
\subsubsection{Imbalanced CIFAR10 and CIFAR100}
The original CIFAR10/CIFAR100 \cite{krizhevsky2009learning} contains 50,000 images for training and 10,000 images for validation with 10/100 categories. Based on the literature \cite{cui2019class,cao2019learning}, two common CIFAR versions, “long-tailed” (LT) and “Step”, with different imbalance degrees in the experiments were used. The "long-tailed" version is generated by changing the number of training samples per class $\dot{n_i}=n_i*\mu^i$, where $i\in(1,C)$ is the class index, $C$ is the total number of classes, $n_i$ is the original number of training images, and $\mu\in(0,1)$. In the "Step" version, the first half of the training set to contains more and the same number of samples (called head classes), and the second half of the class contains fewer and the same number of samples (called tail classes).

In addition, in practical scenarios, not only the imbalance problem is encountered, but also the problem of few samples is often encountered, and these two problems often occur at the same time. To simulate this situation, we construct imbalanced datasets of different imbalance factor (IF) and quantity ratio (QR) by randomly removing samples in each class to comprehensively evaluate how the imbalanced degree of the dataset and the number of samples change the model classification performance. As shown in Figure. \ref{fig4} (a-b), the IF is a measure of the degree of imbalance in the training set. IF=$n_{max}/n_{min}$ is an index proportional to the imbalance of the data distribution, where $n_{max}$ is the number of samples of the most frequent class and $n_{min}$ is the number of samples of the least frequent class in the training set. QR=$n^{'}/n$ represents the proportion of new training set samples to all training samples, where $n$ is the total number of samples in the original training set, and $n^{'}$ is the total number of samples in the new training set after sampling. The research of Cao et al. \cite{cao2019learning} was followed to train the backbone of ResNet-32 \cite{he2009learning,liu2019large} for 200 epochs on a single NVIDIA RTX A4000 GPU.

\subsubsection{ImageNet-LT}
ImageNet-LT \cite{liu2019large} is the subset of ImageNet \cite{deng2009imagenet} and its 
training set contains 115,800 images from 1,000 categories, with a class cardinality ranging from 5 to 1,280. The validation set contains 500 images in each of the classes. To facilitate fair comparisons, the research of Kang et al. \cite{kang2019decoupling} was followed for training the backbone of ResNet-10 on two NVIDIA RTX A4000 GPUs.

\begin{table*}
\renewcommand\arraystretch{2}
\fontsize{9.0pt}{9.0pt}\selectfont
\centering
\caption{Top-1 accuracy $(\%)$ of different Re-weighting methods, Re-sampling methods and loss functions (maximum performance in training repeated ten times).}
\label{table3}
\begin{tabular}{c:c:ccc:ccc:ccc:ccc} 
\hline
\multicolumn{2}{c:}{Dataset}                  & \multicolumn{6}{c:}{Imbalanced CIFAR10}                                                                                                                                                                                 & \multicolumn{6}{c}{Imbalanced CIFAR100}                                                                                                                                                                                 \\ 
\hline\hline
\multicolumn{2}{c:}{Imbalance Type}           & \multicolumn{3}{c:}{long-tailed}                                                                           & \multicolumn{3}{c:}{Step}                                                                                  & \multicolumn{3}{c:}{long-tailed}                                                                           & \multicolumn{3}{c}{Step}                                                                                   \\ 
\hline\hline
\multicolumn{2}{c:}{Imbalance Factor}         & 10                                & 50                                & 100                                & 10                                & 50                                & 100                                & 10                                & 50                                & 100                                & 10                                & 50                                & 100                                \\ 
\hline\hline
\multirow{4}{*}{Weighting} & RW               & 87.2                              & 78.8                              & 72.5                               & 86.1                              & 73.5                              & 68.1                               & 57.7                              & 44.7                              & 39.2                               & 57.1                              & 42.8                              & 39.1                               \\
                           & DRW              & 87.5\textbf{}                     & 80.3\textbf{}                     & 76.3\textbf{}                      & 86.8\textbf{}                     & 75.0\textbf{}                     & 68.7\textbf{}                      & 58.1\textbf{}                     & 45.3\textbf{}                     & 41.5\textbf{}                      & 58.2\textbf{}                     & 44.8\textbf{}                     & 39.9\textbf{}                      \\
                           & cRT-RW           & 87.5                              & 79.7                              & 76.3                               & 86.8                              & 74.8                              & 68.7                               & 58.2                              & 45.1                              & 41.4                               & 58.3                              & 44.5                              & 39.8                               \\
                           & \textbf{PPW}     & \textbf{88.2}                     & \textbf{81.7}                     & \textbf{78.0}                      & \textbf{88.2}                     & \textbf{78.6}                     & \textbf{71.4}                      & \textbf{59.9}                     & \textbf{48.4}                     & \textbf{43.0}                      & \textbf{58.8}                     & \textbf{46.3}                     & \textbf{44.0}                      \\ 
\hline\hline
\multirow{5}{*}{Sampling}  & RS               & 86.5                              & 77.6                              & 72.5                               & 84.7                              & 72.9                              & 64.7                               & 55.9                              & 39.1                              & 34.5                               & 53.5                              & 41.4                              & 38.7                               \\
                           & DRS              & 87.2\textbf{}                     & 79.8\textbf{}                     & 73.2\textbf{}                      & 86.2\textbf{}                     & 73.0\textbf{}                     & 67.1\textbf{}                      & 57.3\textbf{}                     & 45.2\textbf{}                     & 42.0\textbf{}                      & 57.7\textbf{}                     & 44.3\textbf{}                     & 41.2\textbf{}                      \\
                           & cRT-RS           & 87.3                              & 80.0                              & 73.3                               & 86.3                              & 73.1                              & 67.0                               & 57.5                              & 45.4                              & 42.4                               & 58.1                              & 44.7                              & 41.3                               \\
                           & P-B              & 87.0\textbf{}                     & 79.3\textbf{}                     & 73.5\textbf{}                      & 85.9\textbf{}                     & 72.5\textbf{}                     & 64.6\textbf{}                      & 57.5\textbf{}                     & 43.4\textbf{}                     & 40.9\textbf{}                      & 55.2\textbf{}                     & 43.1\textbf{}                     & 39.9\textbf{}                      \\
                           & \textbf{PPS}     & \textbf{87.5}                     & \textbf{80.4}                     & \textbf{74.2}                      & \textbf{86.4}                     & \textbf{73.5}                     & \textbf{67.2}                      & \textbf{57.6}                     & \textbf{45.9}                     & \textbf{42.4}                      & \textbf{58.2}                     & \textbf{44.8}                     & \textbf{41.6}                      \\ 
\hline\hline
\multirow{4}{*}{Loss}      & CE               & 86.7                              & 77.0                              & 70.7                               & 85.9                              & 73.4                              & 67.9                               & 55.7                              & 43.9                              & 38.3                               & 56.0                              & 42.0                              & 39.0                               \\
                           & Focal            & \multicolumn{1}{l}{86.7}          & \multicolumn{1}{l}{77.1}          & \multicolumn{1}{l:}{70.2}          & \multicolumn{1}{l}{83.6}          & \multicolumn{1}{l}{71.8}          & \multicolumn{1}{l:}{63.9}          & \multicolumn{1}{l}{55.8}          & \multicolumn{1}{l}{44.3}          & \multicolumn{1}{l:}{38.4}          & \multicolumn{1}{l}{53.5}          & \multicolumn{1}{l}{41.9}          & \multicolumn{1}{l}{38.6}           \\
                           & LDAM             & \multicolumn{1}{l}{87.0}          & \multicolumn{1}{l}{79.5}          & \multicolumn{1}{l:}{73.4}          & \multicolumn{1}{l}{85.0}          & \multicolumn{1}{l}{73.7}          & \multicolumn{1}{l:}{66.6}          & \multicolumn{1}{l}{56.9}          & \multicolumn{1}{l}{46.0}          & \multicolumn{1}{l:}{39.6}          & \multicolumn{1}{l}{56.3}          & \multicolumn{1}{l}{43.3}          & \multicolumn{1}{l}{39.6}           \\
                           & \textbf{CRI}     & \textbf{87.8}                     & \textbf{79.9}                     & \textbf{75.8}                      & \textbf{87.0}                     & \textbf{74.8}                     & \textbf{68.8}                      & \textbf{58.8}                     & \textbf{46.9}                     & \textbf{41.4}                      & \textbf{56.3}                     & \textbf{44.0}                     & \textbf{41.0}                      \\ 
\hline\hline
\multicolumn{2}{c:}{\textbf{CRI+PPW}}         & 88.9\textbf{}                     & 83.0\textbf{}                     & 79.5\textbf{}                      & 88.7\textbf{}                     & 81.6\textbf{}                     & 77.4\textbf{}                      & 61.0\textbf{}                     & 49.2\textbf{}                     & 44.9\textbf{}                      & 60.1\textbf{}                     & 50.5\textbf{}                     & 47.2\textbf{}                      \\
\multicolumn{2}{c:}{\textbf{CRI+PPW+PPmix}}   & \multicolumn{1}{l}{\textbf{90.6}} & \multicolumn{1}{l}{\textbf{85.8}} & \multicolumn{1}{l:}{\textbf{82.8}} & \multicolumn{1}{l}{\textbf{90.1}} & \multicolumn{1}{l}{\textbf{83.5}} & \multicolumn{1}{l:}{\textbf{80.4}} & \multicolumn{1}{l}{\textbf{62.2}} & \multicolumn{1}{l}{52.5\textbf{}} & \multicolumn{1}{l:}{47.0\textbf{}} & \multicolumn{1}{l}{\textbf{62.2}} & \multicolumn{1}{l}{\textbf{52.3}} & \multicolumn{1}{l}{\textbf{47.4}}  \\
\multicolumn{2}{c:}{\textbf{CRI+PPW+RIDE}}    & \multicolumn{1}{l}{87.5\textbf{}} & \multicolumn{1}{l}{84.2\textbf{}} & \multicolumn{1}{l:}{82.4\textbf{}} & \multicolumn{1}{l}{84.8\textbf{}} & \multicolumn{1}{l}{77.6\textbf{}} & \multicolumn{1}{l:}{74.9\textbf{}} & \multicolumn{1}{l}{60.3\textbf{}} & \multicolumn{1}{l}{\textbf{54.1}} & \multicolumn{1}{l:}{\textbf{50.7}} & \multicolumn{1}{l}{58.9\textbf{}} & \multicolumn{1}{l}{48.3\textbf{}} & \multicolumn{1}{l}{41.9\textbf{}}  \\
\hline
\end{tabular}
\end{table*}

\subsubsection{iNaturalist 2018}
iNaturalist 2018 \cite{van2018inaturalist} is a real-world fine-grained \cite{wei2021fine,zhao2017survey} dataset that is used for classification and detection, consisting of 437,500 images in 8,142 categories, which naturally has an extremely imbalanced distribution. The official distribution of training and validation images was used, and the training of the ResNet-50 backbone followed the research of Kang et al. \cite{kang2019decoupling} on eight NVIDIA RTX A4000 GPUs.

\subsection{Experimental settings}
Following the methods of Zhong et al. \cite{zhang2021bag} and \cite{wang2020long}, the phased progressive learning (PPL) schedule, the coupling-regulation-imbalance (CRI) loss, and their various combinations are introduced. The commonly used top-1 accuracy on Imbalanced CIFAR, ImageNet-LT, and iNaturalist 2018 are used as evaluation metrics. The detailed settings of hyperparameters and training for all datasets are listed in Table \ref{table1}. We conducted experiments on the imbalanced CIFAR datasets to determine the optimal values of $E_0$, $E_1$, and $\rho$ for different IF. It is worth noting that these optimal values vary, as shown in Figure. \ref{fig8} and Figure. \ref{fig9}. However, due to space limitations, we could not include all the values in Table \ref{table1}.

In order to mitigate the significant computational cost resulting from an excessive number of hyperparameters, PPmix empirically uses the best parameters found in PPW as fixed values. To verify the generality of the proposed methods, the training configurations used for the Imbalanced CIFAR datasets are applied directly to other datasets in the hyperparameter optimization process. For example, in ImageNet-LT and iNaturalist 2018, PPW and PPmix are fixed at the power-law form, and $\rho$ is fixed at 5. The phased hyperparameter thresholds are set to $[100,160]$ for 200 epochs of training and $[50,80]$ for 100 epochs of training. In addition, for the CRI loss, according to our experimental results, $\sigma$ is fixed a $\sigma=-(p_y/T)(1-T)^\gamma logT$.

\begin{figure*}
\centering
\includegraphics[scale=0.9]{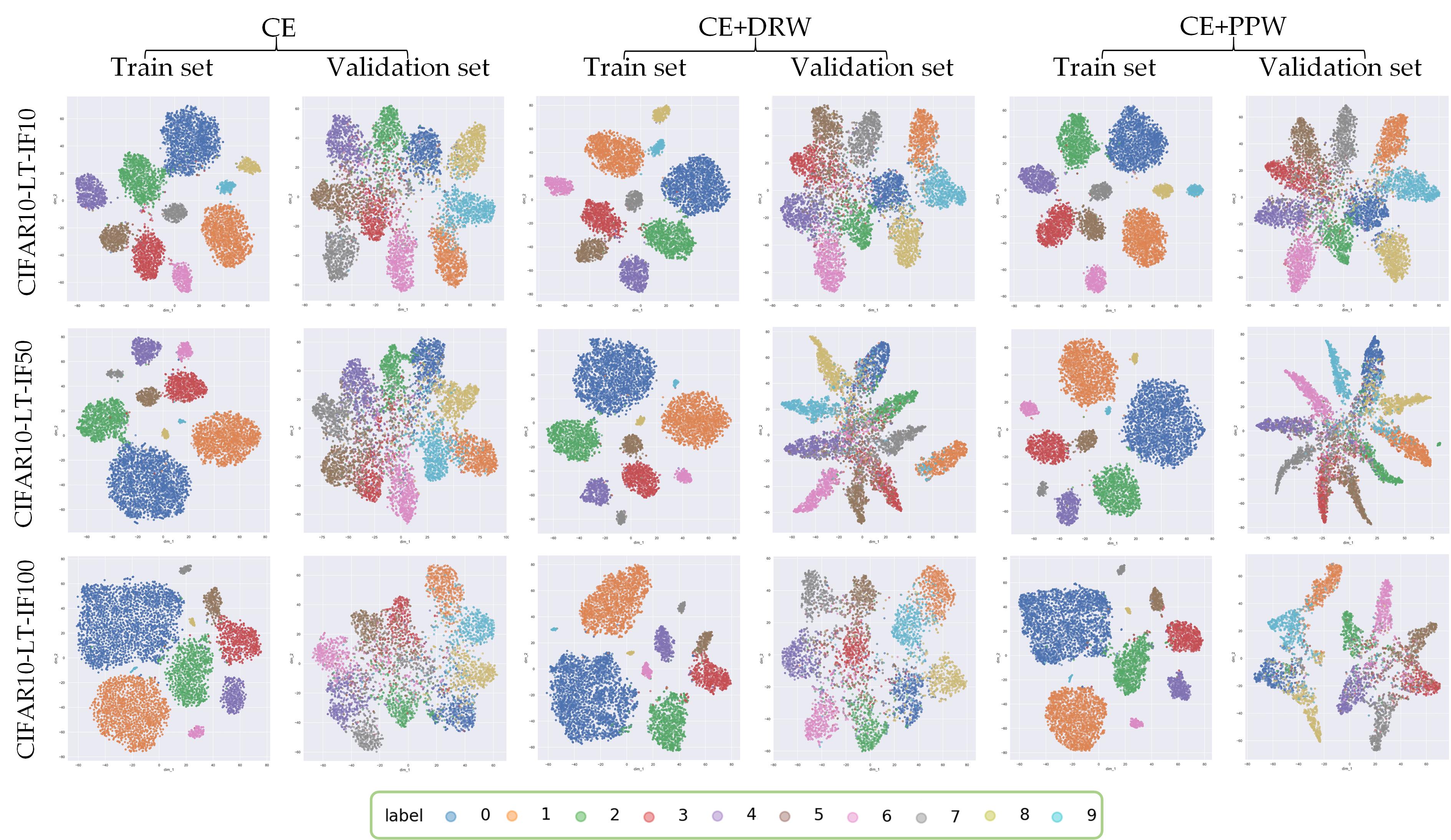}
\caption{t-SNE visualization of the features of the last model layer on CIFAR10-LT (IF=10, 50, and 100). Visualization of models using CE, CE+DRW, and CE+PPW on the training set and validation set.}
\label{fig5}
\end{figure*}

\begin{figure*}
\centering
\includegraphics[scale=0.9]{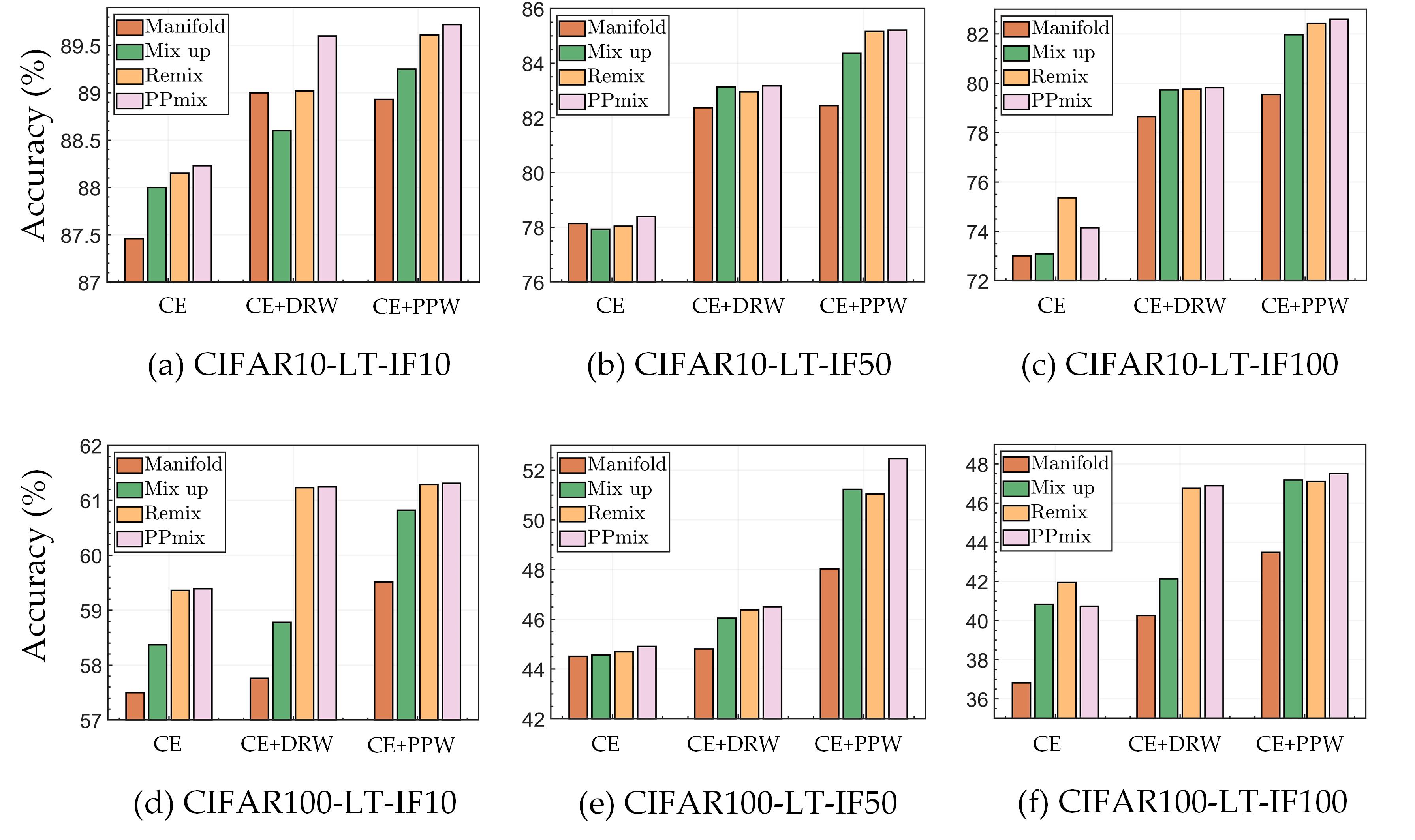}
\caption{The top-1 accuracy of the mixup methods is combined with different re-weighting methods (CE, CE+DRW, and CE+PPW) on CIFAR10-LT and CIFAR100-LT with different IF (10, 50 and 100).}
\label{fig6}
\end{figure*}

\begin{figure*}
\centering
\includegraphics[scale=0.9]{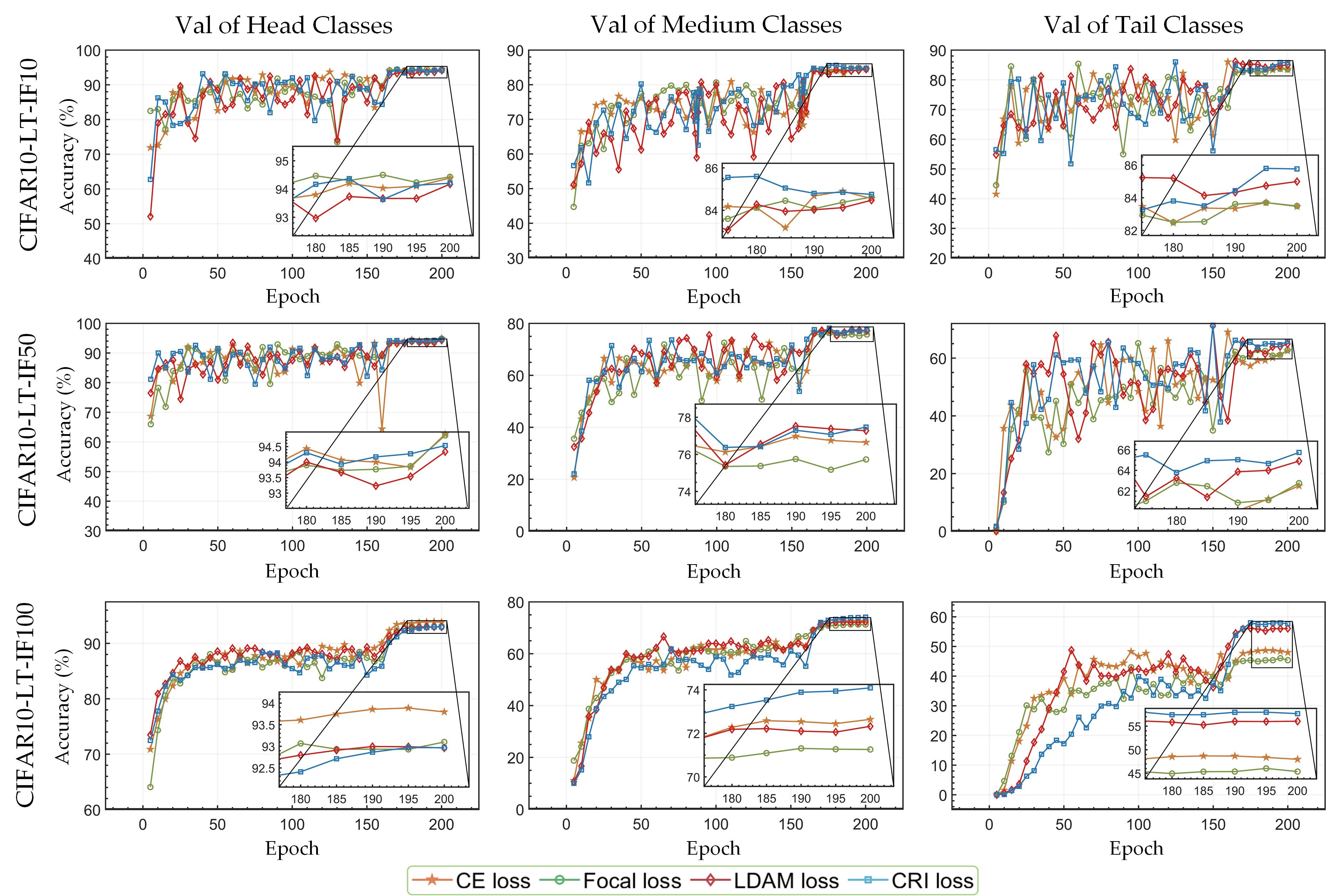}
\caption{Validation accuracy curves under different loss functions on CIFAR10-LT-IF10, CIFAR10-LT-IF50, and CIFAR10-LT-IF100. During the training of 200 epochs, the accuracy of three parts was tested on the validation set for each epoch, especially the most important region of 175-200 epochs, which was enlarged.}
\label{fig7}
\end{figure*}

\begin{table*}
\renewcommand\arraystretch{2}
\fontsize{9.0pt}{9.0pt}\selectfont
\centering
\caption{Top-1 accuracy $(\%)$ on Imbalanced CIFAR10 and CIFAR100 for different architectures (The results of the other methods are all from the original paper).}
\label{table4}
\begin{tabular}{c:ccc:ccc:ccc:ccc} 
\hline
Dataset                 & \multicolumn{6}{c:}{Imbalanced CIFAR10}                                                       & \multicolumn{6}{c}{Imbalanced CIFAR100}                                                        \\ 
\hline\hline
Imbalance Type          & \multicolumn{3}{c:}{Long-tailed}              & \multicolumn{3}{c:}{Step}                     & \multicolumn{3}{c:}{Long-tailed}              & \multicolumn{3}{c}{Step}                       \\ 
\hline\hline
Imbalance Factor        & 10            & 50            & 100           & 10            & 50            & 100           & 10            & 50            & 100           & 10            & 50            & 100            \\ 
\hline\hline
CB-Focal \cite{cui2019class}                & 87.1          & 79.3          & 74.6          & 83.5          & -             & 60.3          & 58.0          & 45.2          & 39.6          & 50.0          & \textbf{-}    & 36.0           \\
LDAM-DRW \cite{cao2019learning}                & 88.2          & 79.3          & 77.0          & 87.8          & -             & 76.9          & 58.7          & -             & 42.0          & 59.5          & -             & 45.4           \\
cRT-mix \cite{kang2019decoupling}                 & 89.8          & 84.2          & 79.1          & -             & -             & -             & 62.1          & 50.9          & 45.1          & -             & -             & -              \\
LWS-mix \cite{kang2019decoupling}                  & 89.6          & 82.6          & 76.3          & -             & -             & -             & 62.3          & 50.7          & 44.2          & -             & -             & -              \\
Remix-DRW \cite{chou2020remix}               & 89.0          & -             & 79.8          & 88.3          & -             & 77.9          & 89.0          & -             & 79.8          & 60.4          & -             & 46.8           \\
Remix-DRS \cite{chou2020remix}               & 88.9          & -             & 79.5          & -             & -             & -             & 60.5          & -             & 46.5          & 60.8          & -             & 47.3           \\
BBN \cite{zhou2020bbn}                     & 88.3          & 82.2          & 79.8          & -             & -             & -             & 59.1          & 47.0          & 42.6          & -             & -             & -              \\
MiSLAS \cite{zhong2021improving}                  & 90.0          & 85.7          & 82.1          & -             & -             & -             & \textbf{63.2} & 52.3          & 47.0          & -             & -             & -              \\
RIDE \cite{wang2020long}                    & -             & -             & -             & -             & -             & -             & -             & -             & 49.1          & -             & -             & -              \\ 
\hline\hline
\textbf{CRI +PPW}       & 88.9          & 83.0          & 79.5          & 88.7          & 81.6          & 77.4          & 61.0          & 49.2          & 44.9          & 60.1          & 50.5          & 47.2           \\
\textbf{CRI +PPW+PPmix} & \textbf{90.6} & \textbf{85.8} & \textbf{82.9} & \textbf{90.1} & \textbf{83.5} & \textbf{80.4} & 62.2          & 52.5          & 47.0          & \textbf{62.2} & \textbf{52.3} & \textbf{47.4}  \\
\textbf{CRI+PPW+RIDE}   & 87.5\textbf{} & 84.2\textbf{} & 82.4\textbf{} & 84.8          & 77.6          & 74.9          & 60.3          & \textbf{54.1} & \textbf{50.7} & 58.9          & 48.3          & 41.9           \\
\hline
\end{tabular}
\end{table*}

\begin{table}
\renewcommand\arraystretch{2}
\fontsize{9.0pt}{9.0pt}\selectfont
\centering
\caption{Top-1 accuracy $(\%)$ on ImageNet-LT (The results of the other methods are all from the original paper).}
\label{table5}
\begin{tabular}{c:c} 
\hline
Dataset                & ImageNet-LT    \\ 
\hline\hline
Backbones              & ResNet-10      \\ 
\hline\hline
CE \cite{cui2019class}                     & 34.0           \\
CB-Focal \cite{cui2019class}              & 32.6           \\
LDAM-DRW \cite{cao2019learning}              & 36.0           \\
Decoupling \cite{kang2019decoupling}            & 41.8           \\
Bag of tricks \cite{zhang2021bag}       & 43.1           \\ 
\hline\hline
\textbf{CRI+PPW+PPmix} & 43.3           \\
\textbf{CRI+PPW+RIDE}  & \textbf{54.9}  \\
\hline
\end{tabular}
\end{table}

\begin{table}
\renewcommand\arraystretch{2}
\fontsize{9.0pt}{9.0pt}\selectfont
\centering
\caption{Top-1 accuracy $(\%)$ on iNaturalist-2018 (The results of the other methods are all from the original paper).}
\label{table6}
\begin{tabular}{c:c} 
\hline
Dataset                & iNaturalist-2018  \\ 
\hline\hline
Backbones              & ResNet-50        \\ 
\hline\hline
LDAM-DRW \cite{cao2019learning}              & 68.0             \\
BBN \cite{zhou2020bbn}                    & 69.6             \\
Remix-DRW \cite{chou2020remix}             & 70.5             \\
MisLAS \cite{zhong2021improving}                 & 71.6             \\
RIDE \cite{wang2020long}                   & 72.6             \\ 
\hline\hline
\textbf{CRI+PPW+PPmix} & 70.5             \\
\textbf{CRI+PPW+RIDE}  & \textbf{72.7}    \\
\hline
\end{tabular}
\end{table}

\begin{figure*}
\centering
\includegraphics[scale=0.9]{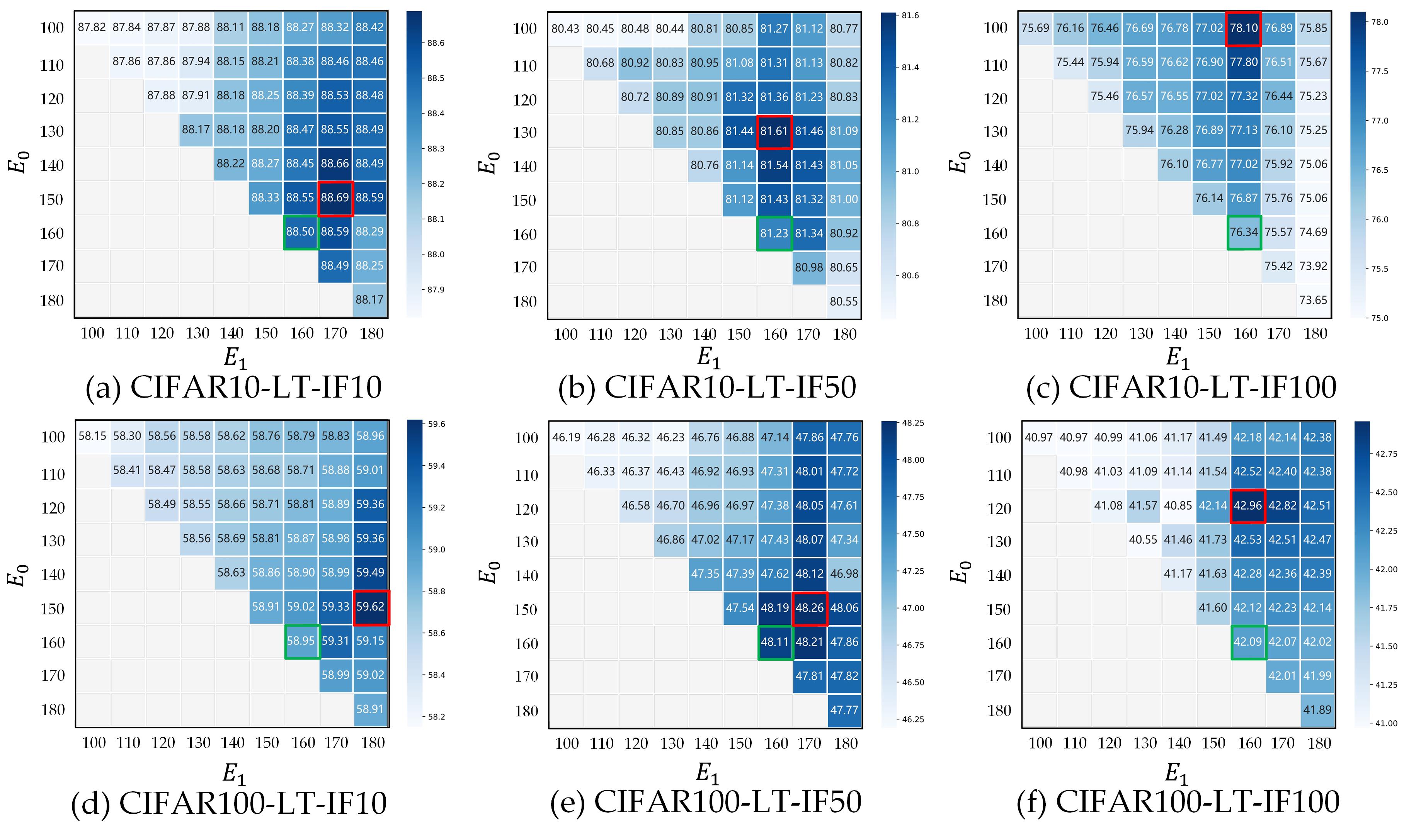}
\caption{Ablation studies of $[E_0, E_1]$ on CIFAR10-LT and CIFAR100-LT, where $f(E)$ is fixed at the power-law form and $\rho=5$.}
\label{fig8}
\end{figure*}

\begin{figure*}
\centering
\includegraphics[scale=0.9]{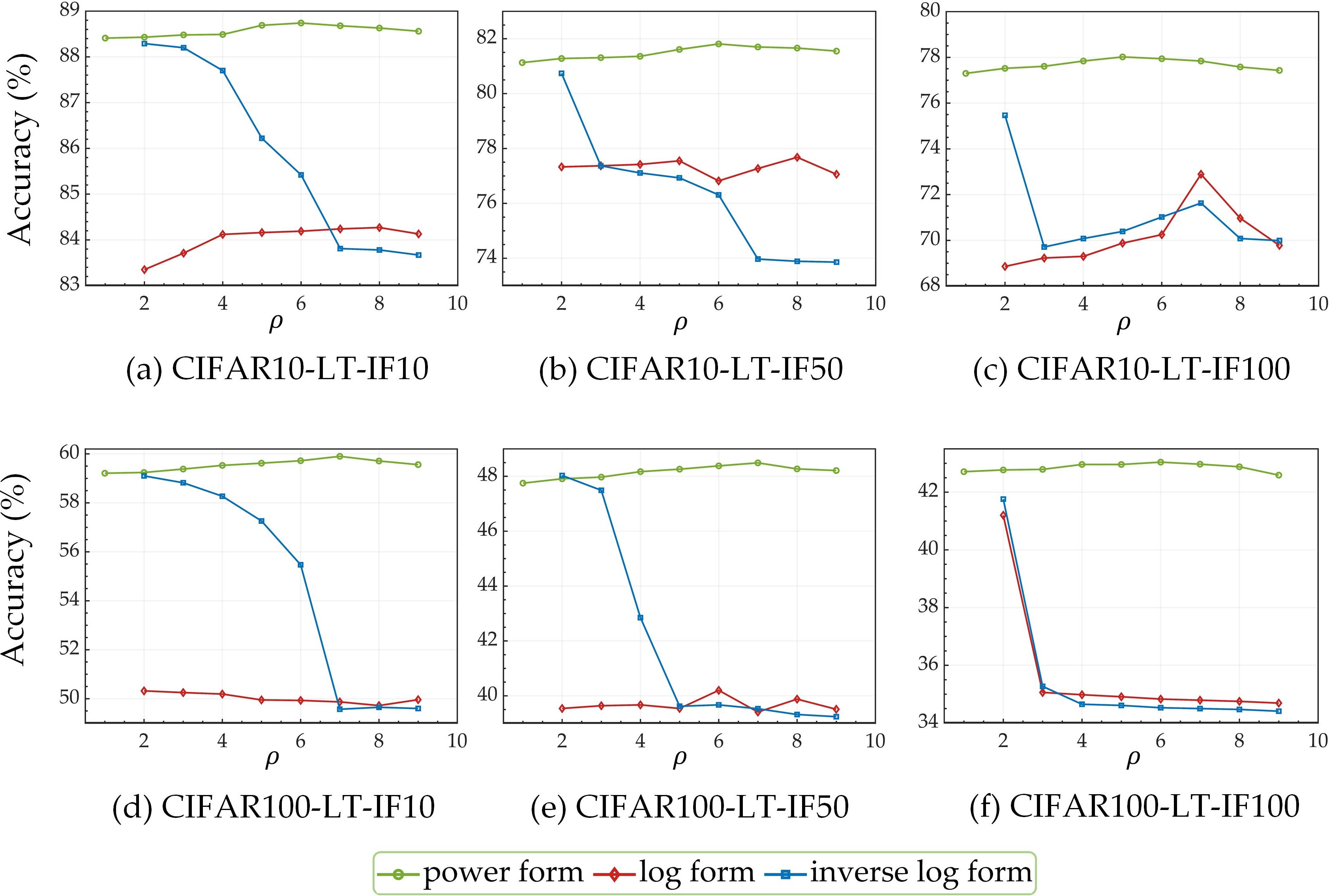}
\caption{Ablation studies of $f(E)$ and $\rho$ on CIFAR10-LT and CIFAR100-LT, the phased training epoch threshold $[E_0, E_1]$ are fixed at [150, 170], [130, 160], [100, 160], [150, 180], [150, 170], and [120,160], respectively.}
\label{fig9}
\end{figure*}

\subsection{Performance test of PPL}
First, we compare the performance of our PPW and PPS methods with RW, RS, DRW, DRS, and cRT under different IF and QR. As shown in Figure. \ref{fig4} (c-f), the PPW, PPS, and existing re-balancing methods were tested on CIFAR10-LT with different IF \cite{cui2019class} and QR. The experimental results (Figure. \ref{fig4} (c) and (e)) show that the accuracy of each method decreases with increasing IF. As IF increases, the performance of the one-stage methods (RW, RS) gradually approaches and eventually exceeds that of the cross-entropy (CE) loss. When IF reaches extreme values (e.g., IF=500), the model will have difficulty converging using the one-stage methods, resulting in a performance that is far worse than that of the CE loss. As the IF increases, the performance advantage of the two-stage approaches (DRW, DRS, cRT) over the CE loss also gradually decreases. However, the PPL methods (PPW, PPS) consistently show the best results, and the performance gap between PPL and other methods increases as the IF increases. Therefore, we can conclude that PPW and PPS can alleviate the problem of dataset bias or domain shift that may be caused by abrupt transitions between stages in two-stage methods, and are more effective when dealing with more extreme imbalanced datasets. In addition, as shown in (Figure. \ref{fig4} (d) and (f)), PPW and PPS outperform all other methods as the QR decreases. Therefore, it is shown that the method of gradually training of the network is also effective for over-fitting caused by repeated training on data sets with insufficient samples.

Similarly, in terms of values, as shown in Table \ref{table2}, when the QR is fixed at 1, the accuracy of PPW at IF=10 is $0.7\%$ better than that of DRW, and the superiority at IF=200 is $2.4\%$. The accuracy gap between PPS and DRS also increases from $0.2\%$ to $1.2\%$. Similarly, when IF is fixed at 100, the accuracy of PPW at QR=1 is $1.7\%$ better than that of DRW, and the superiority reaches $3.5\%$ at QR=0.5. The accuracy gap between PPS and DRS also increases from $0.9\%$ to $1.1\%$. As a result, the PPW and PPS have greater adaptability and robustness, especially when dealing with more extreme imbalances and smaller datasets.

Next, we extend our analysis to the imbalanced CIFAR datasets with different IF and Step versions. Table \ref{table3} shows the best performances of different re-balancing methods, including common one-stage methods (RW, RS), two-stage approaches (DRW, DRS, cRT), the phased progressive weighting (PPW), and the phased progressive sampling (PPS). The models used in this study are trained using CE loss. Our experimental results show that PPL achieves remarkable improvements on CIFAR datasets with varying factors. It is worth noting that the PPS method differs from the basic extension of the progressively-balanced (P-B) sampling. It involves the addition of critical initial and final stages during data training. Here, the initial phase provides appropriate initial parameters for subsequent training, and the final phase continuously contributes to a self-adaptive classifier. The PPS method has been shown to provide a performance improvement of $0.1-3\%$ over the conventional P-B approach.

It should be noted that we also trained the datasets using re-weighting and re-sampling simultaneously, but our results indicate that there is no discernible advantage to using both techniques simultaneously over using either in isolation. As a result, we have found that instead of using both techniques simultaneously, it is optimal to use them separately. Furthermore, the performance of the PPW method exceeds that of the PPS as shown in Table \ref{table3}. Therefore, the PPW method is adopted as the baseline in all subsequent experiments.

As shown in (Figure. \ref{fig5}), to further compare the performance of the different weighting methods, the features of the last model layer on the training set and the validation set of CIFAR10-LT are visualized. It is noteworthy that all four methods produce very clear class boundaries regardless of the degree of imbalance of the training set. However, as IF increases, PPW produces clearer class boundaries than CE and DRW on the validation set, which means better class separation. 

In addition, phased progressive mixup (PPmix), Mix up \cite{zhang2017mixup}, Manifold mixup \cite{verma2019manifold}, and Remix \cite{chou2020remix} are tested based on the CE loss on the Imbalanced CIFAR datasets. The performance of the mixup methods is further tested in combination with RW, DRW, and PPW. As can be seen in (Figure. \ref{fig6}), PPmix alone does not perform particularly well, but it outperforms Remix when used in combination with DRW or PPW. At the same time, the PPW used in this study performs significantly better than DRW when combining different mixing methods, and PPmix+PPW performs best.

\subsection{Performance test of CRI loss}
The second part of Table \ref{table3} shows the top-1 validation accuracy of models using different loss functions on the original CIFAR-10 and CIFAR-100 datasets. Only different loss functions are used during the training process instead of a combination with RW, RS, etc. methods. It can be observed that the proposed coupling-regulation-imbalance (CRI) loss performs better than the CE loss, Focal loss, and LDAM loss. These results confirm the effectiveness of improving performance by addressing the classification difficulty imbalance and mitigating the resulting loss of outliers.

To further demonstrate the generality of the CRI loss, we evaluate the performance of the head classes (1,200+ images per class), medium classes (200-1,200 images per class), and tail classes (less than 200 images per class) of CIFAR10-LT-IF10, CIFAR10-LT-IF50, and CIFAR10-LT-IF100. As shown in Figure. \ref{fig7}, compared to models using the CE loss on CIFAR10-LT-IF10, although the accuracy of the head classes decreases by $0.3\%$, the performance of the CRI loss on the medium and tail classes improves by $0.2\%$ and $2.2\%$, respectively. Similarly, for CIFAR10-LT-IF50, the performance of the CRI loss decreases by $0.3\%$ in the head classes, but increases by $0.8\%$ and $3.2\%$ in the medium and tail classes compared to the CE loss. For CIFAR10-LT-IF100, although it decreases by $0.7\%$ on the head classes, the performance of the CRI loss on the medium and tail classes increases by more than $1.4\%$ and $10\%$ compared to the CE loss. In addition, LDAM loss and Focal loss perform similarly to the CRI loss in the head classes, but worse in the medium and tail classes.

As mentioned above, both the CRI loss and the PPW performance are the best compared to other similar methods, so the combination of the two methods is used in the following experiments. First, the performance of CRI+PPW is tested. Then the proposed regularization PPmix is introduced (denoted as CRI+PPW+PPmix), and the performance further improved significantly. In addition, to mitigate the problem of a decrease in the accuracy of head classes under CRI loss, CRI+PPW is applied in the routing of diverse distribution-aware experts (RIDE) \cite{wang2020long}, which is denoted as CRI +PPW+RIDE. As seen in Table \ref{table3}, CRI+PPW performs better than pure PPW, and CRI+PPW+PPmix performs better than all previous results. CRI +PPW+RIDE works best on CIFAR100-LT-IF50 and CIFAR100-LT-IF100. It can be seen that the proposed PPL method and other regularization methods such as RIDE can also be well combined with our CRI loss.

\subsection{Comparing our methods with other state-of-the-art methods}
\subsubsection{Experimental results on Imbalance CIFAR}
To verify the efficiency of the proposed method, methods including CB-Focal \cite{cui2019class}, LDAM-DRW \cite{cao2019learning}, cRT-mix \cite{kang2019decoupling}, LWS-mix \cite{kang2019decoupling}, Remix-DRW \cite{chou2020remix}, BBN \cite{zhou2020bbn}, MisLAS \cite{zhong2021improving}, and RIDE \cite{wang2020long} are also used for comparative validation. The results are listed in Table \ref{table4} and show that CRI+PPW+PPmix performs the best on all versions of CIFAR10-LT, CIFAR10-Step, and CIFAR100-Step. For CIFAR100-LT, CRI+PPW+PPmix outperforms all previous methods at IF=50, and is only worse than RIDE at IF=100 and MiSLAS at IF=10. CRI+PPW+RIDE has the best results at IF=100 and IF=50 for CIFAR100-LT, but its performance is worse than CRI+PPW+PPmix for CIFAR 10-LT.

\subsubsection{Experimental results on large-scale imbalanced datasets }
The effectiveness of the methods used in this study will be further verified on two large-scale imbalanced datasets, ImageNet-LT and iNaturalist 2018 are further verified. Table \ref{table5} and Table \ref{table6} show the experimental results on ImageNet-LT and iNaturalist 2018, respectively.  The CRI+PPW+PPmix method outperforms the previous best Bag of tricks \cite{zhang2021bag} by $0.2\%$, and the CRI +PPW+RIDE further further by $11\%$ On ImageNet-LT. On iNaturalist 2018, the CRI+PPW+RIDE also beats the previous best RIDE by $0.1\%$.

\subsection{Ablation study}
Some hyperparameters in the proposed PPL need to be optimized: the phased training epoch threshold $[E_0, E_1]$ and the progressive hyperparameter $\rho$. Taking the training PPW as an example, due to the huge cost and time spent on ImageNet-LT and iNaturalist 2018, we choose to conduct experiments on CIFAR10-LT and CIFAR100-LT. From epoch 100 to epoch 180, both $E_0$ and $E_1$ are changed in the same interval, and the learning rate (LR) drop-in occurs at epoch 160 and 180. The performance matrix is shown in (Figure. \ref{fig8}) $(E_0\leq E_1)$. When $E_0=E_1$, the progressive transition phase is canceled and the PPW degenerates to the DRW. The traditional DRW method after annealing the LR only plays a minor role in the backpropagation of the front layers. At the same time, the depth feature update is small and the overall model cannot better fit the imbalanced dataset. Taking CIFAR10-LT-IF100 (Figure. \ref{fig9} (c)) as an example, the accuracy is further improved by $1.73\%$ compared to conventional DRW ($E_0$=$E_1$=160, green square) when $E_0$=100 and $E_1$=160 (red square) in progressive training. Since LR decreases at epoch 160 and the progressive training starts at epoch 100, backpropagation is not too weak, and can better fit the imbalanced datasets.

Figure. \ref{fig9} also shows the performance of CIFAR10-LT and CIFAR100-LT under three different forms of the transformation function $f(E)$ (power-law form, log form, and inverse log form) with different progressive hyperparameters $\rho$. Taking the training CIFAR10-LT-IF100 as an example (Figure. \ref{fig9} (c)), the data show that the power-law form with $\rho=5$ is more effective.

\section{Conclution}
In this paper, two methods are proposed: phased progressive learning (PPL) schedule and coupling-regulation-imbalance (CRI) loss. To alleviate the problem of data bias or domain shift that is caused by two-stage approaches, PPL adopts a smooth transition from the general pattern of representation learning to classifier training, thereby facilitating classifier learning without harming the representation learning of the network. The larger imbalances or fewer samples the datasets are, the more effective PPL will be. At the same time, CRI loss can more effectively deal with the problem of quantity imbalance, limiting huge losses from outliers and keeping the focus-of-attention on different classification difficulties. The methods in this paper have served to improve performance on various benchmark vision tasks, can be nested in other methods, and we will further develop our method for specific object detection and semantic segmentation tasks in the future.

\bibliographystyle{elsarticle-num}
\bibliography{cas-refs}

\bio{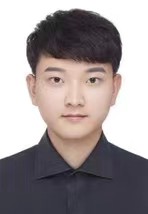}
Liang Xu received the BS degree from North China Electric Power University in 2020. He is currently pursuing the Ph.D. degree at the University of Science and Technology of China (USTC), and his current research interests include deep learning and multi-organ intelligent interaction.
\endbio

\vskip3pt
\vskip3pt

\bio{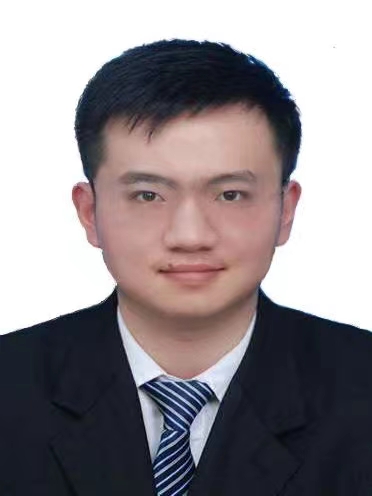}
Yi Cheng received the BS degree from Hefei University of Technology in 2021. He is currently working on his master's degree at USTC. His current research interests include deep learning and intelligent detection of circulating tumors based on microfluidics.
\endbio

\vskip3pt
\vskip3pt
\vskip3pt
\vskip3pt
\vskip3pt

\bio{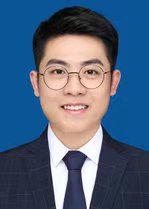}
Fan Zhang received the BS degree from Dalian University of Technology in 2018. He is currently pursuing the PhD degree at USTC. His current research interests include remote surgical navigation and intelligent medical diagnosis.
\endbio

\vskip3pt
\vskip3pt
\vskip3pt
\vskip3pt
\vskip3pt
\vskip3pt
\vskip3pt
\vskip3pt
\vskip3pt

\bio{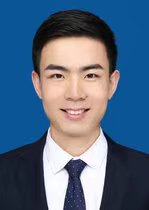}
Bingxuan Wu received his BS degree from USTC in 2018. He is currently pursuing a PhD degree at USTC. His current research interests include remote surgical navigation and medical image processing based on multimodality.
\endbio

\vskip3pt
\vskip3pt
\vskip3pt
\vskip3pt
\vskip3pt
\vskip3pt
\vskip3pt
\vskip3pt
\vskip3pt
\vskip3pt

\bio{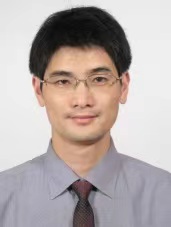}
Pengfei Shao received his Ph.D. from USTC in 2000. He has been an associate professor at USTC since 2001. His research interests include medical device development, multimodal biomedical imaging, and image navigation therapy.
\endbio

\vskip3pt
\vskip3pt
\vskip3pt
\vskip3pt
\vskip3pt
\vskip3pt
\vskip3pt
\vskip3pt

\bio{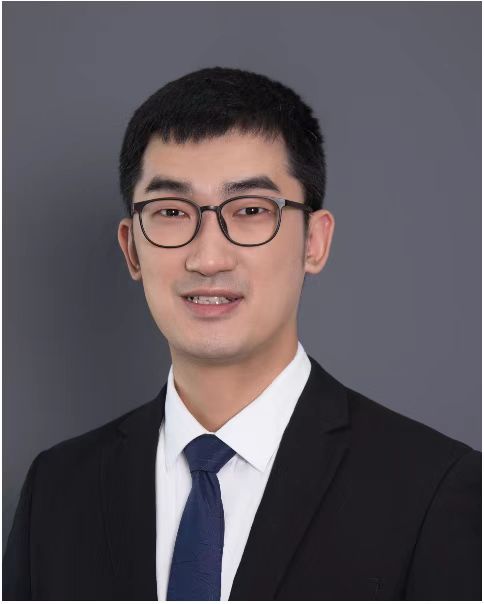}
Peng Liu received his Ph.D. degree from USTC in 2016. He is currently a special associate researcher at the Suzhou Institute for Advanced Research, USTC. His research interests include multimodal medical imaging technology and surgical navigation technology.
\endbio

\vskip3pt
\vskip3pt
\vskip3pt
\vskip3pt

\bio{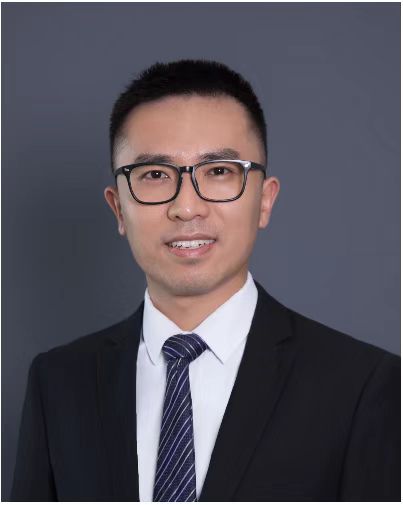}
Shuwei Shen received his Ph.D. from USTC in 2019. He is currently a special associate researcher at the Suzhou Institute for Advanced Research, USTC. His research interests include tissue optical phantom preparation and AI-based early screening of medical diseases.
\endbio

\vskip3pt
\vskip3pt
\vskip3pt
\vskip3pt
\vskip3pt
\vskip3pt
\vskip3pt

\bio{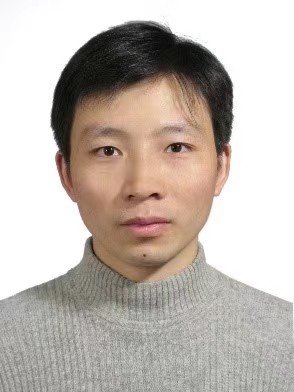}
Peng Yao received his Ph.D. from USTC in 2005. From 2014 to 2016, he was an academic visiting scholar at the Chinese University of Hong Kong. His main research directions are biometric recognition, medical image processing, industrialization of iris recognition technology.
\endbio

\vskip3pt
\vskip3pt
\vskip3pt
\vskip3pt
\vskip3pt
\vskip3pt

\bio{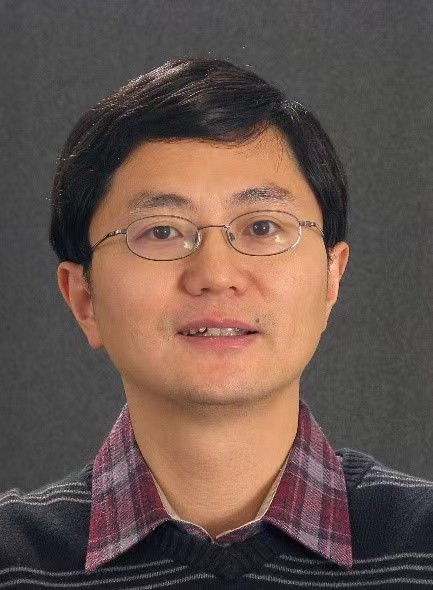}
Ronald X. Xu received his Ph.D. from the Massachusetts Institute of Technology in 1999. He worked as a tenured associate professor at Ohio State University. He is currently a professor at the Suzhou Institute for Advanced Research, USTC. He is a member of the Institute of Physics and a senior member of the Society of Photo-Optical Instrumentation Engineers (SPIE). His research interests include artificial intelligence and medical diagnosis, micronano drug packaging. He has conducted more than 20 research projects and published more than 100 scientific papers in high-impact SCI journals. His research has been featured in Columbus CEO magazine and he has been named one of Ohio's top ten people of the year and two of the biggest stars in scientific research. He has received the Wallace H. Coulter Young Achievement Award in Translational Medicine, the Ohio TechColumbus Inventor of the Year Award, and the Lumbley Research Award.
\endbio

\end{document}